\documentclass{article}

\usepackage{PRIMEarxiv}

\usepackage{authblk}
\usepackage[utf8]{inputenc} 
\usepackage[T1]{fontenc}    
\usepackage{url}            
\usepackage{booktabs}       
\usepackage{amsfonts}       
\usepackage{nicefrac}       
\usepackage{microtype}      
\usepackage{fancyhdr}       
\usepackage{graphicx}       

\usepackage[hidelinks]{hyperref}       
\usepackage{amsmath}
\usepackage[table,xcdraw]{xcolor}
\usepackage{listings}
\usepackage{array}
\usepackage{float}
\usepackage{algorithm}
\usepackage{algpseudocode}

\usepackage{multirow}
\usepackage{rotating}

\usepackage{caption}
\usepackage{subcaption}
\usepackage{verbatim}

\usepackage[toc,page]{appendix}
\usepackage{tikz}
\usetikzlibrary{shapes.geometric, arrows.meta, positioning, calc, fit}
\usepackage{pgfplots}
\usepackage{svg}
\usepackage{bbm}

\graphicspath{{media/}}     

\title{DENOGRAD: A GRADIENT-BASED FRAMEWORK FOR DATA REFINEMENT IN TABULAR AND TIME-SERIES LEARNING}

\author[ ]{
 J. Javier Alonso-Ramos \textsuperscript{1,2},
 Ignacio Aguilera-Martos \textsuperscript{1,2},
 Francisco Herrera \textsuperscript{1,2},
 Andr\'es Herrera-Poyatos \textsuperscript{3}
 }

\affil[1]{Department of Computer Science and Artificial Intelligence, University of Granada, Granada, Spain.}
\affil[2]{Andalusian Institute of Data Science and Computational Intelligence (DaSCI), University of Granada, Spain.}
\affil[ ]{\texttt{Emails: \{jjalonso, nacheteam, andreshp\}@ugr.es, herrera@decsai.ugr.es}}

\pgfplotsset{compat=1.18}
\begin{document}
\maketitle

\begin{abstract}
In the Data-Centric Artificial Intelligence (AI) paradigm, improving data quality is essential for robust machine learning. However, many denoising methods rely on rigid statistical assumptions or require clean reference data, which limits their applicability in real-world scenarios. In this work, we propose DenoGrad, a gradient-based framework for data refinement that leverages a pretrained neural network to iteratively correct noisy observations by optimizing the input space while keeping the model fixed.
DenoGrad is applicable to both tabular regression and time-series forecasting, and incorporates a consensus-based strategy to ensure temporally coherent updates in sequential settings. Experiments on ten real-world datasets show that the proposed approach yields consistent improvements in downstream predictive performance while preserving the statistical structure of the data, as measured by distributional and correlation-based metrics. In addition, DenoGrad can improve generalization in nominally clean datasets, acting as a form of dataset-level regularization. These results support model-guided data refinement as a practical component of data-centric machine learning workflows.
Code is available at: \url{https://github.com/ari-dasci/S-DenoGrad}.

\end{abstract}

\section{Introduction}
\label{sec:introduction}

In the emerging paradigm of Data-Centric Artificial Intelligence (AI), training-data fidelity is a central determinant of success. Although model architectures continue to improve, recent surveys show that data quality across training and production is often the main bottleneck for predictive performance \cite{zha2025data}. As a result, noise reduction has become a standard preprocessing step. Yet, especially in classification and regression, most approaches still rely on \textit{filtering}, that is, identifying and removing noisy instances \cite{Zhang2026, Szeghalmy2024}, rather than \textit{correcting} them \cite{songLearningNoisyLabels2023}. While filtering can be effective for high-confidence noise, it inevitably causes data loss and may remove difficult but valid samples, reducing dataset size and diversity \cite{northcuttPervasiveLabelErrors}. Traditional corrective denoisers face the opposite risk: overcorrection, which can distort the original distribution and harm downstream interpretability \cite{Mokari2025ACS}.

Noise strongly impairs a model's ability to approximate the underlying data distribution \cite{Santhosh2023ImpactOA}. It distorts decision boundaries and regression manifolds, pushing models to memorize stochastic errors instead of learning generalizable patterns. This is especially damaging for classical methods or limited-capacity models, where signal and noise become hard to separate, leading to poor generalization and unstable predictions \cite{zhangUnderstandingDeepLearning2021}.

By contrast, Deep Learning (DL) architectures show a distinctive resilience. Empirical and theoretical studies indicate that neural networks, especially when trained with Stochastic Gradient Descent, can learn useful representations even under Gaussian noise \cite{camuto2020explicit}. This robustness arises from the optimization dynamics: early in training, gradients are dominated by the true signal, while zero-mean random noise tends to cancel out across iterations \cite{liuEarlyLearningRegularizationPrevents2020}. Although later training can overfit noise, the loss landscape learned in this early phase still captures a strong prior of the true data manifold \cite{arpitCloserLookMemorization2017}. This creates an opportunity: the same ability that helps DL models separate signal from noise during training can be used to refine the data itself through gradient information.

In this paper, we present DenoGrad, a gradient-based denoiser designed to improve the performance of arbitrary machine learning models. Instead of defining noise through fixed statistical heuristics, we define it through a learned model, arguing that the semantic knowledge encoded in a neural network is better suited to correct instances than rigid assumptions. DenoGrad is model-independent: it freezes a pre-trained neural backbone and iteratively backpropagates corrections into the input space. In this sense, it can be viewed as a data-centric reverse of adversarial optimization. Whereas adversarial attacks perturb inputs to maximize prediction error, DenoGrad minimizes the loss with respect to the input features to move corrupted observations toward the clean manifold learned by the network. This iterative adjustment, related to feature visualization and adversarial image inversion \cite{mahendranUnderstandingDeepImage2015}, acts as a universal preprocessing step for any subsequent regressor. Unlike knowledge distillation, DenoGrad operates directly on the dataset, using the implicit knowledge of a frozen backbone to produce a refined version of the data.

To validate DenoGrad, we designed an experimental pipeline over ten diverse datasets (five tabular and five time-series) corrupted with varying levels of Gaussian noise. We compare it against seven established statistical and learning-based denoisers, and evaluate the recovered data across downstream architectures ranging from Ridge, kNN, and XGBoost to TabPFN and xLSTM.

The main contributions of this work are summarized as follows:

\begin{itemize}
    \item \textbf{A Novel Gradient-Based Denoising Framework:} We introduce DenoGrad, a model-independent method that optimizes inputs to correct noisy features without clean reference data or secondary training phases.
    \item \textbf{Optimal Performance-Fidelity Trade-off:} Through rigorous statistical testing (Critical Difference (CD) diagrams), we show that DenoGrad delivers consistent predictive gains while preserving the topology of the original data, achieving the lowest Sliced Wasserstein distance (SWD) \cite{kolouri2019generalized} and the highest feature correlation consistency ($\bar\rho$) among the evaluated methods.
    \item \textbf{Universal Downstream Generalization:} Unlike standard filters which depend strongly on the downstream model, we show that the manifold recovered by DenoGrad consistently benefits any subsequent regressor.
    \item \textbf{Dataset-Level Regularization and Latent Noise Mitigation:} We show that DenoGrad yields substantial predictive gains even without synthetic corruption. By harmonizing small inter-variable inconsistencies and suppressing latent aleatory noise in real-world data, it acts as a dataset-level regularizer and extends denoising into practical data refinement.
\end{itemize}

While DenoGrad is designed as a general-purpose data refinement framework, it is primarily expected to be effective when noise is stochastic and approximately zero-mean (e.g., Gaussian). It may be less suitable in settings dominated by strong systematic biases, as discussed in Section~\ref{sec:discussion;subsec:limitations}.

To interpret the empirical results, we distinguish three related but conceptually different effects: (i) \textit{noise correction}, which removes stochastic perturbations from corrupted observations; (ii) \textit{data refinement}, which aligns samples with the learned manifold to improve structural coherence; and (iii) \textit{dataset-level regularization}, which reduces small inter-variable inconsistencies even in nominally clean data. Depending on the regime, DenoGrad may exhibit all three.

The remainder of this paper is organized as follows. Section \ref{sec:background} reviews existing denoising strategies and the DL dynamics that support our approach. Section \ref{sec:gradient-based-denoiser} presents DenoGrad, including the mathematical formulation of the input optimization process and the consensus strategy for sequential data. Section \ref{sec:experiment} describes the experimental methodology, datasets, metrics, and baselines. Section \ref{sec:results} analyzes the results, including statistical benchmarking, hyperparameter ablations, and computational efficiency. Section \ref{sec:discussion} discusses the theoretical boundaries and practical guidelines of DenoGrad. Finally, Section \ref{sec:conclusions} summarizes the main findings and future research directions.


\section{Background and Related Work}
\label{sec:background}

This section provides the theoretical foundation of DenoGrad. We describe the challenges of noise in regression and time-series settings, review existing denoising methods, and summarize the deep learning dynamics that enable self-supervised noise reduction through input optimization.

\subsection{Noise in Regression and Time Series Analysis}
\label{sec:background;subsec:noise}

In data analysis, noise is any unwanted signal modification that obscures underlying information \cite{garcia2014dealing}. Although much work has focused on \textit{label noise} in classification \cite{garcia2019enabling}, here we address \textit{attribute} and \textit{target noise} in regression, where corruption appears in continuous feature spaces \cite{Johnson2022ASO}. Regression noise is commonly modeled as $y_{obs}=f(x)+\epsilon$, with $\epsilon \sim N(0,\sigma^2)$ \cite{kwak2017central}. In practice, data often departs from this ideal through systematic bias, input-dependent heteroscedasticity, and impulsive outliers \cite{shankar2020three, Bishop1996RegressionWI}.

Noise is especially critical in time-series forecasting. Unlike independent tabular samples, time series contain temporal dependencies, so discarding noisy points can break sequence structure, distort lag features, and bias trend estimation. In autoregressive settings, noise at time $t$ also propagates to future predictions, amplifying errors over long horizons \cite{barnett2024generative}. Effective preprocessing for sequential data should therefore \textit{correct}, not \textit{discard}, corrupted observations.

\subsection{State-of-the-Art Denoising Strategies}
\label{sec:background;subsec:denoising}

Denoising seeks to recover the clean latent signal $x$ from a noisy observation $x^*$, and is a core component of inverse problems and machine learning pipelines \cite{Milanfar2024DenoisingAP}. The methods considered in our experiments fall into three families: time-domain filters, spectral decompositions, and learning-based reconstructions.

Time-domain and statistical filters operate directly on observed sequences. Moving Average (MA) \cite{shan2022novel} smooths short-term fluctuations but introduces lag and blurs sharp transitions. For linear systems with Gaussian noise, the Kalman Filter (KF) \cite{10443058} yields optimal recursive estimates, but often struggles under strongly non-linear dynamics. In multivariate tabular settings, Principal Component Analysis (PCA) \cite{9729514} projects samples onto orthogonal components and removes low-variance noise, assuming that noise is orthogonal to the main signal subspace.

For non-stationary signals, spectral and frequency-domain decompositions are common. Wavelet Transform Decomposition (WTD) separates signals into time-frequency scales and thresholds high-frequency coefficients. It is powerful and widely used, including in deep learning hybrids \cite{Li2022dnswin}, but depends heavily on choosing a fixed mother wavelet. Empirical Mode Decomposition (EMD) \cite{boudraa2006denoising} instead adaptively decomposes signals into Intrinsic Mode Functions (IMFs). This avoids fixed bases but is computationally expensive and prone to mode-mixing artifacts.

Deep learning denoisers currently define the state of the art (SOTA) by mapping noisy inputs to clean outputs with non-linear architectures. Denoising Autoencoders (DAE) \cite{Cui2025} remove noise by reconstructing inputs through a low-dimensional bottleneck, and this principle has been extended to variational frameworks guided by physical imaging models. More recently, Denoising Residual Networks (DN-ResNet) \cite{Huang2025} use skip connections and transformer-based components to estimate noise directly.

Despite strong performance, most learning-based methods are strictly supervised and require paired clean-noisy training data ($X_{clean},X_{noisy}$), which is often unavailable. More broadly, all three families have key constraints: statistical and spectral methods rely on rigid \textit{a priori} assumptions (e.g., pure Gaussianity or high-frequency dominance), while supervised deep models depend on clean labels. DenoGrad addresses this gap by reversing the paradigm and letting the predictive model implicitly define noise structure in a self-supervised, data-driven way.

\subsection{Deep Learning Dynamics: Spectral Bias}
\label{sec:background;subsec:dl}

To justify using a model trained on noisy data as a denoiser, we rely on Spectral Bias in Deep Neural Networks (DNNs) \cite{rahaman2019spectral}. DNNs tend to learn low-frequency, generalizable structure before fitting high-frequency stochastic variations. Arpit et al. \cite{arpitCloserLookMemorization2017} further showed that true-signal examples are learned early, while memorization of complex or noisy instances appears later in training.

Therefore, even when trained on noisy data, a model can still encode the dominant structure of the underlying manifold and approximate the clean signal. DenoGrad explicitly exploits this property to guide data refinement.

\subsection{Gradient-Based Input Optimization and Data Refinement}
\label{sec:background;subsec:gradient_opt}

Once a predictive backbone captures the dominant data manifold, we can invert the usual optimization direction: freeze network weights $\theta$ and backpropagate error gradients directly into the input space $X$ and $Y$.

Gradient-based input optimization is well established in interpretability and representation analysis. Saliency maps \cite{adebayo2018sanity}, feature visualization, activation maximization \cite{nguyen2016synthesizing}, and image inversion \cite{mahendranUnderstandingDeepImage2015} refine or reconstruct inputs by optimizing activations or reconstruction losses. These methods show that trained networks encode rich manifold structure that can be used to iteratively refine inputs via gradient descent.

Under this view, DenoGrad is an inverse adversarial optimization process. While attacks such as Fast Gradient Signed Method \cite{goodfellow2014explaining} perturb inputs to maximize error ($x_{adv}=x+\eta \cdot \nabla_xL$), DenoGrad applies the opposite update ($x_{denoised}=x-\eta \cdot \nabla_xL$). Iterative descent in input space pushes corrupted observations toward regions of higher predictive consistency.

Similar gradient-driven ideas appear in Score-Based Generative Models and Diffusion Models \cite{song2019generative, croitoru2023diffusion}, which estimate score functions to progressively remove noise. Despite strong results in perceptual domains, they require explicit full-distribution modeling and high computational cost. This overhead limits practicality for tabular regression and time-series forecasting, where preserving statistical feature relationships is essential \cite{kotelnikov2023tabddpm}.

Outside generative modeling, dataset refinement has been widely explored in Learning with Noisy Labels. Methods such as Joint Optimization \cite{tanaka2018joint}, PENCIL \cite{yi2019probabilistic}, and co-teaching \cite{zheng2021meta} treat labels as learnable variables updated during training. However, they are intrinsically classification-oriented: they depend on discrete distributions and cross-entropy objectives, so adaptation to continuous regression is not straightforward.

DenoGrad bridges these lines of work by adapting gradient-based input optimization to a purely \textit{discriminative} setting. By exploiting spectral bias in standard predictive backbones (e.g., MLPs, CNNs, or LSTMs), it provides a lightweight, model-independent mechanism to refine noisy tabular and sequential data without complex generative modeling or clean reference targets.

\section{DenoGrad: A Gradient-based Denoiser Framework}
\label{sec:gradient-based-denoiser}

DenoGrad mitigates noise in both input features and target variables by using gradient information from a DL predictive model. Instead of training a separate denoiser, it reuses the downstream predictive model as a correction guide, making the approach lightweight and easy to integrate into existing machine learning pipelines.

The framework is model-independent and can operate with any differentiable predictive model, but its effectiveness depends on the representational quality of the chosen backbone. In other words, DenoGrad is architecture-agnostic, not performance-invariant. This coupling is also an advantage: as DL backbones improve at modeling complex manifolds, DenoGrad receives more informative gradients and improves denoising performance without structural changes.

The core premise of DenoGrad relies on \textit{spectral bias}: during training, neural networks learn dominant low-frequency structure (signal) before overfitting high-frequency irregularities (noise) \cite{arpitCloserLookMemorization2017}. Thus, even when trained only on noisy data, a sufficiently regularized model can capture a smoothed approximation of the underlying manifold, therefore no clean ground truth is required. Although gradients do not guarantee convergence to the exact clean manifold, they are typically oriented toward low-frequency structure and are effective for correcting noisy instances.

Building on this property, DenoGrad reverses standard training through \textit{Input Optimization}. In conventional training, model parameters $\theta$ are updated to fit data. Here, the trained model is frozen and data instances are treated as trainable variables, iteratively adjusted to align with the learned manifold. The predictive model acts as an oracle that provides correction directions via backpropagated gradients. Unlike knowledge distillation, which transfers behavior between models, DenoGrad operates only in data space: it optimizes features and targets against a fixed oracle to output a refined dataset, not a new model.

\begin{figure}
    \centering
    \begin{subfigure}[b]{0.47\textwidth}
        \centering
        \includegraphics[width=0.8\textwidth]{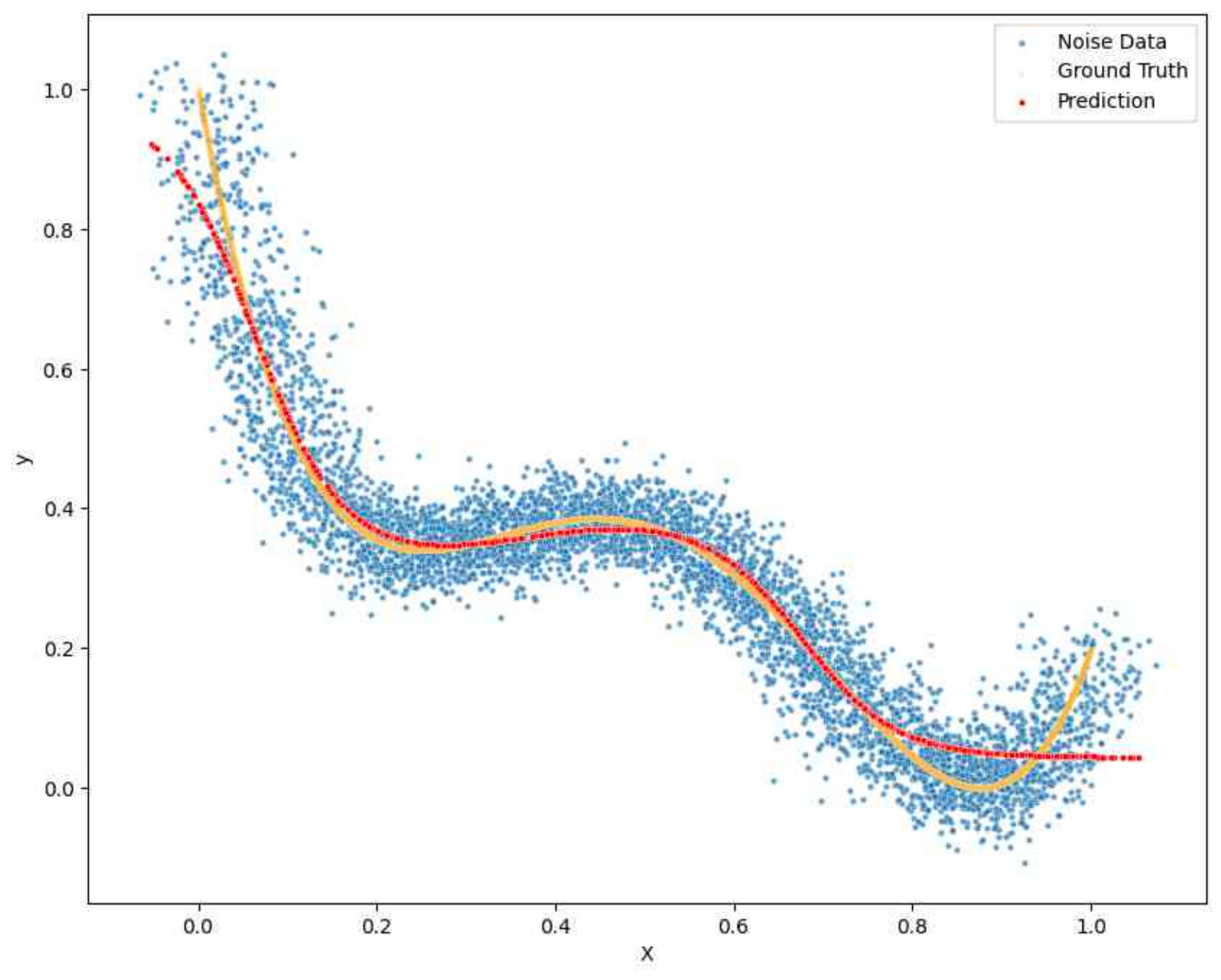}
        \caption{DL model (red) fitted to noisy data (blue) alongside the ground truth (orange).}
        \label{fig:2d_example_a}
    \end{subfigure}
    \hfill
    \begin{subfigure}[b]{0.47\textwidth}
        \centering
        \includegraphics[width=0.8\textwidth]{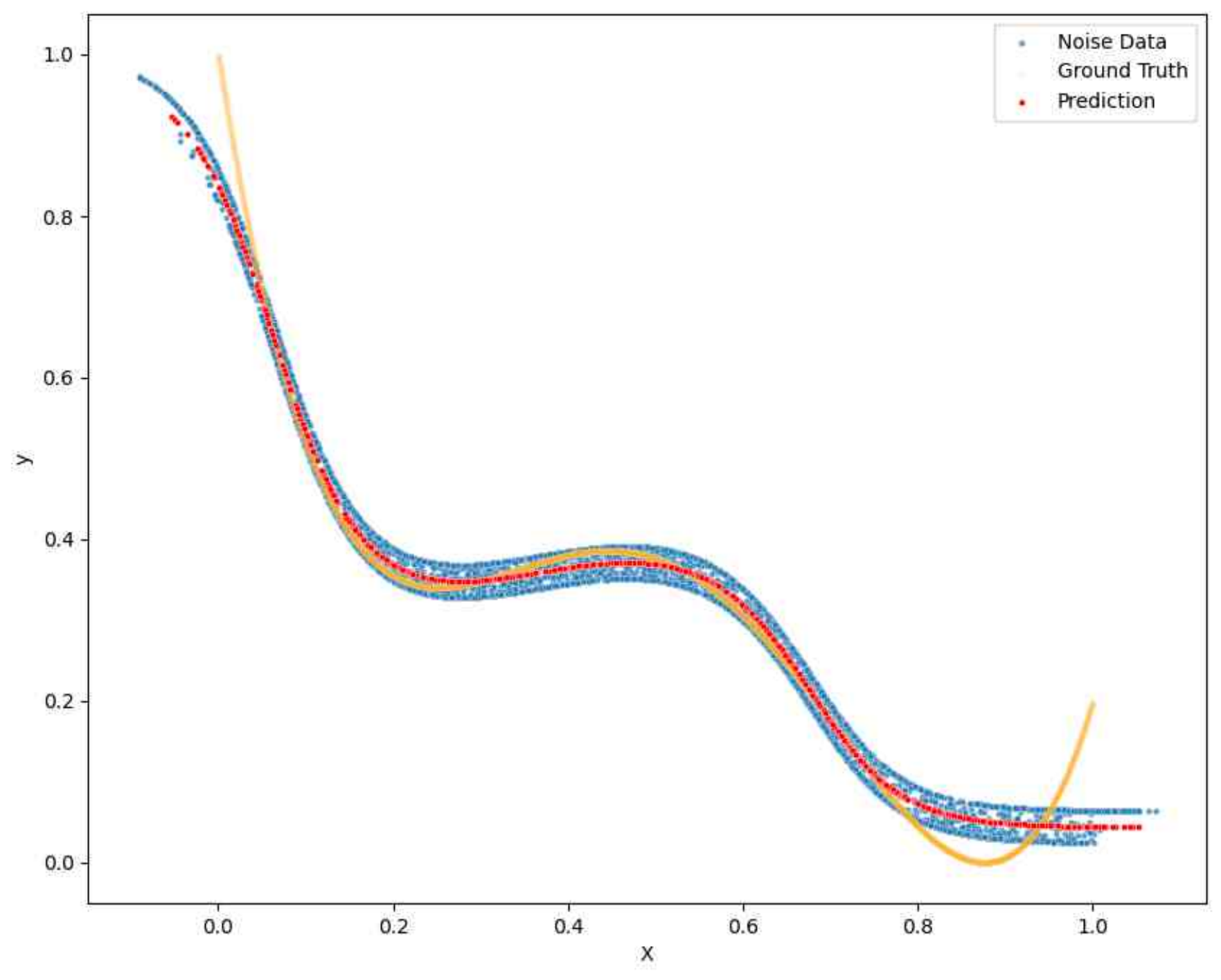}
        \caption{After DenoGrad: samples are shifted toward the learned manifold.}
        \label{fig:2d_example_b}
    \end{subfigure}
    \caption{2D noisy data before and after applying DenoGrad.}
    \label{fig:2d_example}
\end{figure}

This alignment is illustrated in Figure \ref{fig:2d_example}. A regression model trained on noisy data captures the dominant trend, and the learned function acts as an attractor: after DenoGrad, samples move toward the model manifold, reducing variance while preserving global structure.

A key practical advantage is production efficiency. Unlike DL denoisers that train a separate cleaning model, DenoGrad reuses the same predictive model used downstream. It also applies a holistic correction across all dimensions, refining both features and targets while preserving inter-variable structure.

The framework has been released as an open-source Python library to ensure reproducibility and facilitate its integration into third-party pipelines. It is available for installation via the Python Package Index (PyPI)\footnote{Installation command: \texttt{pip install denograd}}, and the complete source code, including documentation and usage examples, is hosted on GitHub\footnote{\url{https://github.com/ari-dasci/S-DenoGrad}}.

The remainder of this section presents the formal mathematical definition (Section~\ref{sec:formalization}) and the time-series extension via a consensus strategy (Section~\ref{sec:time-series}). The empirical rationale for jointly refining features and targets is analyzed in Section~\ref{sec:results;subsec:ablation_denoise_y}.

\subsection{Mathematical Formulation}
\label{sec:formalization}

We now provide a formal description of DenoGrad in three parts: problem definition, optimization objective with hyperparameters, and gradient-guided refinement.

\subsubsection*{Problem Formulation.}
Let $\mathcal{D}_{noisy} = \{(x_i, y_i)\}_{i=1}^N$ be an observed dataset, where each instance is a noisy realization of an underlying clean sample $(x_i^*, y_i^*)$, i.e., $x_i = x_i^* + \epsilon_{x,i}$ and $y_i = y_i^* + \epsilon_{y,i}$, with stochastic noise terms $\epsilon_{x,i}$ and $\epsilon_{y,i}$. DenoGrad refines $(x_i, y_i)$ toward a consistent approximation of the latent clean manifold learned by a predictive model $f_\theta$ trained on $\mathcal{D}_{noisy}$.

\subsubsection*{Objective and Hyperparameters.}
Refinement is formulated as loss minimization with respect to data instances while keeping model parameters $\theta$ fixed. In regression, we use mean squared error (MSE): $\mathcal{L}(f_\theta(x), y) = \|f_\theta(x) - y\|_2^2$.

The optimization is governed by two key hyperparameters:
\begin{itemize}
    \item Noise Reduction Rate (NRR-$\eta$): controls the magnitude of each update step in feature space.
    \item Tolerance Threshold ($\tau$): defines a gating condition that prevents updates when the prediction error satisfies $|f_\theta(x) - y| \le \tau$, preserving high-confidence samples and preventing over-smoothing.
\end{itemize}

The gating mechanism is encoded by a binary mask $\mathbbm{I}_{noisy}$ that activates updates only for instances deemed noisy. By preserving controlled stochasticity in low-error samples, it acts as implicit regularization, avoids collapse to an artificially smooth manifold, and improves downstream generalization \cite{bishop1995training}.

\subsubsection*{Gradient-based Refinement.}
DenoGrad computes the sensitivity of the loss with respect to both inputs and targets as
$g_x = \nabla_{x} \mathcal{L}(f_\theta(x), y), \quad g_y = \nabla_{y} \mathcal{L}(f_\theta(x), y)$.

To balance corrections across heterogeneous variables, gradients are jointly normalized by concatenating them and dividing by their $L_2$ norm: $\hat{g}_x = \frac{g_x}{\|[g_x, g_y]\|_2 + \epsilon}, \quad
\hat{g}_y = \frac{g_y}{\|[g_x, g_y]\|_2 + \epsilon}$.

This \textit{Joint Normalization} scales updates to local loss-landscape steepness and prevents one variable from dominating the correction. The iterative refinement step is $x \leftarrow x - \eta \cdot \hat{g}_x \cdot \mathbbm{I}_{noisy}, \quad
y \leftarrow y - \eta \cdot \hat{g}_y \cdot \mathbbm{I}_{noisy}$.

\begin{figure}
    \centering
    \includegraphics[width=0.8\linewidth]{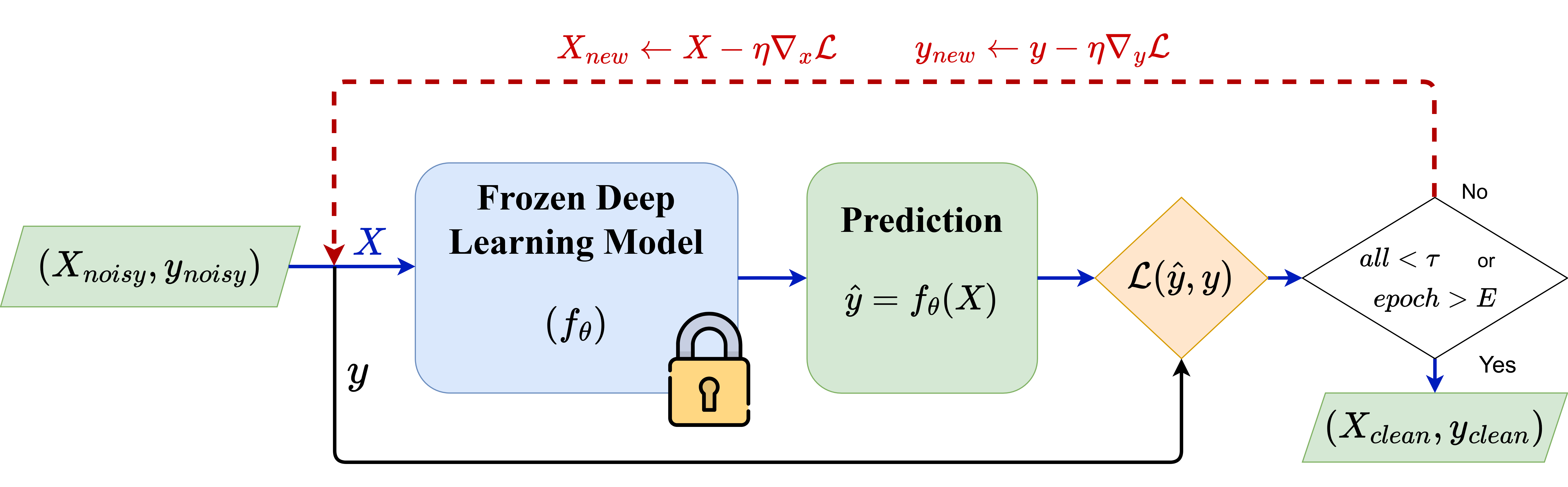}
    \caption{The DenoGrad Pipeline: The model remains frozen while the loss is backpropagated to the input space, iteratively refining the features and targets. The optimization loop terminates either upon reaching the maximum number of epochs or when all instance errors are reduced below the predefined threshold $\tau$.}
    \label{fig:denograd_pipeline}
\end{figure}

In practice, this is implemented iteratively over the dataset: gradients are computed by backpropagation and applied directly in input space while model parameters remain frozen. A schematic overview appears in Figure~\ref{fig:denograd_pipeline}.

\subsection{Time-Series Extension: Consensus Strategy}
\label{sec:time-series}

Applying DenoGrad to time series introduces a key challenge: temporal dependency. Unlike independent tabular instances, a point at time $t$, denoted $x_t$, appears in multiple overlapping sliding windows. For window size $W$, $x_t$ can be the last element of the window starting at $t-W+1$, the first element of the window starting at $t$, and any intermediate position. A naive tabular-style update would therefore create conflicts, because gradients from different windows may suggest different corrections for the same $x_t$.

\begin{figure}
    \centering
    \includegraphics[width=0.45\linewidth]{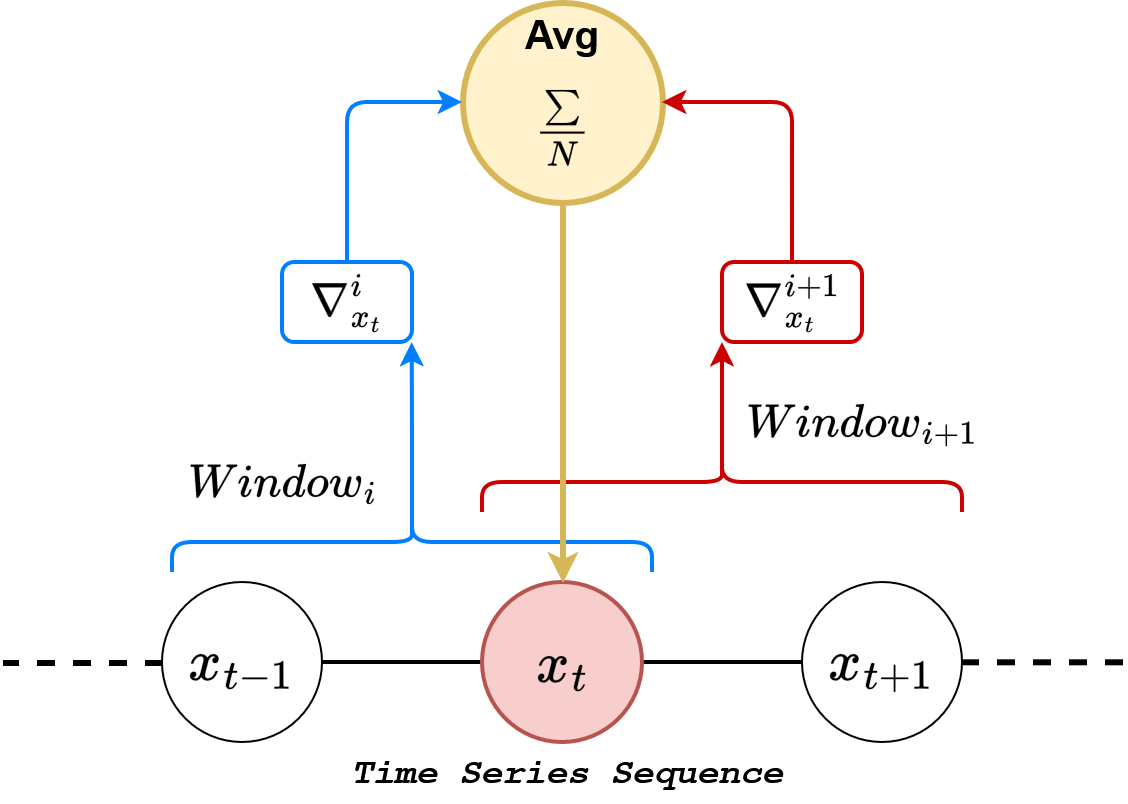}
    \caption{Time-Series Consensus Strategy. A data point $x_t$ (highlighted in red) participates in multiple overlapping windows (e.g., Window $i$ and $i+1$). Gradients computed in each context are accumulated and averaged to produce a single, temporally consistent update vector for that time step.}
    \label{fig:ts_consensus}
\end{figure}

To solve this, DenoGrad uses a \textit{Consensus Strategy via Gradient Accumulation} (Figure~\ref{fig:ts_consensus}). Rather than updating after each batch, it accumulates proposed corrections for each time step across all covering windows. Let $G_t$ be the accumulated gradient for time step $t$, and $C_t$ the number of contributing windows.

At each epoch, the framework traverses all sliding windows, computes gradients, and accumulates them into global buffers at matching time indices. After the full pass, the update for time step $t$ is obtained by averaging accumulated gradients as $x_t^{new} = x_t^{old} - \eta \cdot \frac{G_t}{C_t}$.

This mechanism ensures temporal consistency: each update reflects a consensus direction across all contexts where the observation appears. The same logic applies to target $y$, allowing DenoGrad to denoise both features and targets in sequential settings. When using recurrent models such as LSTMs or GRUs, hidden-state computational graphs must be retained to preserve temporal signal flow during denoising.

\section{Experimental Set-up}
\label{sec:experiment}

We conducted a comprehensive experimental study to evaluate DenoGrad on both tabular regression and time-series forecasting across ten real-world datasets. To assess robustness, we benchmark DenoGrad against seven SOTA denoising methods introduced in Section \ref{sec:background;subsec:denoising}.

Rather than relying on a single predictor, we evaluate denoising effects across a broad set of downstream regressors. This helps ensure that improvements come from better data quality, not from the inductive bias of one architecture.

All experiments were run on a workstation with an AMD Ryzen 9 5950X CPU, an NVIDIA RTX 3070 Ti (8GB VRAM), and 64GB RAM, using Ubuntu 22.04, PyTorch 2.5, and CUDA 12.4.

\subsection{Evaluation Strategy}
\label{sec:experiment;subsec:protocol}

To assess DenoGrad rigorously, we use a comparative protocol that evaluates both downstream predictive gain and preservation of the data manifold.

For predictive evaluation, we compare standard deployment performance with denoised performance. Let $\mathcal{D}_{train}$ and $\mathcal{D}_{test}$ be the original noisy train/test splits. Training and evaluating a downstream regressor on these raw splits yields baseline error ($\mathrm{MSE}_{base}$). We then compare it with denoised counterparts, $\hat{\mathcal{D}}_{train}$ and $\hat{\mathcal{D}}_{test}$, produced by DenoGrad or competing methods. If the model trained on refined data achieves lower error ($\mathrm{MSE}_{deno} < \mathrm{MSE}_{base}$), denoising is successful. Structural integrity (manifold preservation) is evaluated separately through statistical divergences.

To ensure statistical reliability and proper hyperparameter tuning, all datasets were split into training (70\%), validation (10\%), and testing (20\%) sets. For time series, splits were strictly sequential to preserve temporal dependencies. For tabular data, samples were shuffled before splitting.

To keep injected-noise impact comparable across variables with different scales, all datasets were standardized before experimentation. Each continuous feature and target was scaled to zero mean and unit variance ($\mu=0, \sigma=1$). For the main benchmark, we injected synthetic Gaussian noise with fixed standard deviation $\sigma = 0.1$ (Section \ref{sec:results;subsec:sota}). Robustness to higher noise intensities is analyzed separately (Section \ref{sec:results;subsec:noise}).

\subsection{Metrics}
\label{sec:experiment;subsec:metrics}

We use a dual evaluation strategy to capture both \textit{predictive gain} and \textit{distributional integrity} after denoising.

To quantify performance improvement, we use Mean Squared Error ($\mathrm{MSE}$) and Percentage Improvement ($\mathrm{Imp}\%$) of denoised error relative to baseline. A positive $\mathrm{Imp}\%$ indicates successful noise mitigation: $\mathrm{MSE} = \frac{1}{n} \sum_{i=1}^{n} (y_i - \hat{y}_i)^2$, $\mathrm{Imp}\% = \left( \frac{\mathrm{MSE}_{base} - \mathrm{MSE}_{deno}}{\mathrm{MSE}_{base}} \right) \times 100$.


For data integrity, denoising must preserve statistical structure. We therefore measure distributional shift between noisy input distribution $P$ and denoised output distribution $Q$ with $\mathrm{SWD}$. Compared with information-theoretic measures such as Kullback-Leibler divergence, which can be numerically unstable under disjoint empirical supports and do not capture feature-space geometry well, Wasserstein-based metrics provide a more robust topological view. Exact Wasserstein distance is computationally intractable in high dimensions, so $\mathrm{SWD}$ uses random linear projections to reduce the problem to 1D optimal transport solved efficiently by sorting. This yields a mathematically rigorous and tractable measure of manifold preservation \cite{kolouri2019generalized}. $\mathrm{SWD}$ is defined as $\mathrm{SWD}(P, Q) = \int_{\mathbb{S}^{d-1}} W_2(\theta_{\#} P, \theta_{\#} Q) \, d\theta$, where $\mathbb{S}^{d-1}$ represents the $d$-dimensional unit sphere, $\theta_{\#} P$ denotes the one-dimensional marginal projection of distribution $P$ onto direction $\theta$, and $W_2$ is the standard 1D Wasserstein-2 distance. \footnote{In our empirical evaluation, the SWD is computed using the Python Optimal Transport (POT) library \cite{flamary2021pot}, approximating the continuous integral over the unit sphere via 100 random linear projections.}

Additionally, we measure $\bar{\rho} = \frac{1}{M} \sum_{j=1}^{M} \text{corr}(X_{*,j}, \hat{X}_{*,j})$, the average Pearson correlation between each feature column in noisy and denoised datasets. Values near 1 indicate preserved signal shape and trends.

\subsection{Datasets}
\label{sec:experiment;subsec:datasets}

The study uses ten real-world datasets spanning finance, physics, IoT, and healthcare, so the framework is tested across varied modalities and complexity levels. Five are static tabular datasets and five are time-series datasets. Their main characteristics are summarized in Table \ref{tab:datasets}.

\begin{table}[htb]
\renewcommand{\arraystretch}{1.2}
\begin{minipage}[t]{0.35\linewidth}
\centering
\footnotesize
\captionof{table}{Summary of the datasets. Frequencies: \textbf{D}aily, \textbf{H}ourly.}
\label{tab:datasets}
\begin{tabular}{lllrrl}
\hline
\textbf{Name} & \textbf{Type} & \textbf{Src} & \textbf{Inst.} & \textbf{F.} & \textbf{Freq.} \\ \hline
\rowcolor[HTML]{EFEFEF} House Prices \cite{House_Sales_in_King_County_USA} & Tab. & Real & 21,436 & 19 & - \\
Lattice Ph. \cite{lattice-physics_pwr_fuel_assembly_neutronics_simulation_results_1091} & Tab. & Real & 24,000 & 40 & - \\
\rowcolor[HTML]{EFEFEF} Parkinsons \cite{parkinsons_telemonitoring_189} & Tab. & Real & 5,875 & 20 & - \\
RT IOT \cite{rt-iot2022__942} & Tab. & Real & 117,915 & 82 & - \\
\rowcolor[HTML]{EFEFEF} Support2 \cite{support2} & Tab. & Real & 8,579 & 33 & - \\ \hline
Daily Clim. \cite{daily_climate} & TS & Real & 1,576 & 4 & D \\
\rowcolor[HTML]{EFEFEF} ECL \cite{haoyietal-informer-2021} & TS & Real & 6,000 & 320 & H \\
ETT \cite{haoyietal-informer-2021} & TS & Real & 17,420 & 7 & H \\
\rowcolor[HTML]{EFEFEF} MS Stock \cite{microsoft_stock} & TS & Real & 2,192 & 5 & D \\
WTH \cite{haoyietal-informer-2021} & TS & Real & 35,064 & 12 & H \\ \hline
\end{tabular}
\end{minipage}
\hfill
\begin{minipage}[t]{0.47\linewidth}
\centering
\footnotesize
\captionof{table}{Hyperparameters for Downstream Models.}
\label{tab:downstream_params}
\resizebox{0.8\textwidth}{!}{%
\begin{tabular}{llp{2.6cm}}
    \hline
    \textbf{Domain} & \textbf{Model} & \textbf{Key Settings} \\ \hline
    \rowcolor[HTML]{EFEFEF}
    \cellcolor{white} & Ridge & $\alpha=1.0$ \\
    \cellcolor{white} & kNN & $k=5$, uniform \\
    \rowcolor[HTML]{EFEFEF}
    \cellcolor{white} & XGBoost & n=100, depth=6, lr=0.1 \\
    \cellcolor{white} & Dense NN & [64,32,16], ReLU, Adam \\
    \rowcolor[HTML]{EFEFEF}
    \cellcolor{white}\multirow{-5}{*}{\textbf{Tabular}}
                      & TabPFN & N\_ensemble=32 \\ \hline
    \cellcolor{white} & XGBoost & Window features, n=100 \\
    \rowcolor[HTML]{EFEFEF}
    \cellcolor{white} & LSTM & Hidden=50, L=1, Drop=0.2 \\
    \cellcolor{white} & xLSTM & mLSTM, Hidden=32 \\
    \rowcolor[HTML]{EFEFEF}
    \cellcolor{white} & CNN-LSTM & Conv1D(64,k=3) $\to$ LSTM(50) \\
    \cellcolor{white}\multirow{-5}{*}{\textbf{TS}}
                      & DLinear & Linear Trend/Seasonal \\ \hline
\end{tabular}
}
\end{minipage}
\end{table}

\subsection{Comparison Baselines and Downstream Models}
\label{sec:experiment;subsec:models}

To validate DenoGrad, we benchmark it against established denoisers from the three families introduced earlier. For time-domain/statistical filters, we include KF, MA, and PCA. For spectral/frequency-domain methods, we include EMD and WTD. For DL-based denoising, we compare against DAE and DN-ResNet. 

To verify that denoising efficacy generalizes across inductive biases, we use a diverse set of downstream regressors. For tabular data, we evaluate Ridge Regression \cite{hoerl1970ridge} (linear baseline), kNN \cite{cover1967nearest} (instance-based), XGBoost \cite{chen2016xgboost} (gradient boosting trees), a Dense Neural Network (MLP) \cite{lecun2015deep}, and TabPFN \cite{hollmann2023tabpfn} transformer.

For time-series data, we evaluate models that capture temporal dependencies through different mechanisms: XGBoost (with sliding-window adaptation) \cite{chen2016xgboost}, LSTM \cite{hochreiter1997long}, xLSTM \cite{beck2024xlstm}, CNN-LSTM \cite{donahue2015long}, and DLinear \cite{zeng2023transformers}. Hyperparameters and architectural settings for all downstream models are summarized in Table \ref{tab:downstream_params}.

\section{Results and Analysis}
\label{sec:results}

This section presents a comprehensive evaluation of DenoGrad. We first compare against SOTA denoisers to establish empirical competitiveness. We then validate joint feature-target optimization through ablation, and finally analyze behavior across downstream regressors, backbone architectures, hyperparameter settings, noise levels, and computational trade-offs.

\subsection{Comparison with State-of-the-Art Denoisers}
\label{sec:results;subsec:sota}

The primary objective of a denoising framework is to improve downstream predictive accuracy without damaging the underlying data manifold. Table \ref{tab:main_results} reports Average Percentage Improvement (Imp\%), SWD, and $\bar{\rho}$ across all benchmark tabular and time-series datasets.

\begin{table*}
\centering
\caption{Comprehensive evaluation of denoising methods across all datasets. For each dataset, we report the Imp\% $\uparrow$, $\mathrm{SWD} \downarrow$, and $\bar{\rho} \uparrow$. Best results per metric are highlighted in \textbf{bold}.}
\label{tab:main_results}
\resizebox{0.8\textwidth}{!}{%
\begin{tabular}{ll|cccccccc}
\toprule
\textbf{Dataset} & \textbf{Metric} & \textbf{\textbf{DenoGrad (Ours)}} & \textbf{DAE} & \textbf{DN-ResNet} & \textbf{PCA} & \textbf{WTD} & \textbf{EMD} & \textbf{KF} & \textbf{MA} \\
\midrule
\multicolumn{10}{c}{\textit{Static Tabular Datasets}} \\
\midrule
\multirow{3}{*}{House Prices} & Imp\% $\uparrow$ & 77.1\% & 76.2\% & 17.6\% & 89.9\% & 64.0\% & -83.2\% & \textbf{98.5\%} & 64.1\% \\
 & $\mathrm{SWD} \downarrow$ & \textbf{0.0068} & 0.0746 & 0.0137 & 0.0486 & 0.6630 & 0.3252 & 0.7967 & 0.5432 \\
 & $\bar{\rho} \uparrow$ & \textbf{0.9994} & 0.9845 & 0.9990 & 0.9775 & 0.5324 & 0.5510 & 0.2677 & 0.4715 \\
\midrule
\multirow{3}{*}{Lattice Physics} & Imp\% $\uparrow$ & 48.5\% & 83.8\% & 36.0\% & 62.2\% & \textbf{99.6\%} & -552.5\% & 99.1\% & 76.8\% \\
 & $\mathrm{SWD} \downarrow$ & \textbf{0.0019} & 0.3321 & 0.0166 & 0.0344 & 0.8800 & 0.4570 & 0.8424 & 0.5563 \\
 & $\bar{\rho} \uparrow$ & \textbf{1.0000} & 0.8126 & 0.9997 & 0.9757 & 0.1247 & 0.5311 & 0.2184 & 0.4467 \\
\midrule
\multirow{3}{*}{Parkinsons} & Imp\% $\uparrow$ & 72.2\% & 53.8\% & 21.6\% & 61.2\% & 70.9\% & 49.6\% & \textbf{82.8\%} & 63.2\% \\
 & $\mathrm{SWD} \downarrow$ & \textbf{0.0060} & 0.0907 & 0.0270 & 0.0674 & 0.2179 & 0.1411 & 0.3113 & 0.2076 \\
 & $\bar{\rho} \uparrow$ & 0.9981 & 0.9828 & \textbf{0.9985} & 0.9784 & 0.8676 & 0.8426 & 0.7456 & 0.8284 \\
\midrule
\multirow{3}{*}{RT IOT 2022} & Imp\% $\uparrow$ & 82.0\% & 76.3\% & 52.1\% & 57.1\% & 24.2\% & -93414.3\% & \textbf{98.0\%} & 84.3\% \\
 & $\mathrm{SWD} \downarrow$ & \textbf{0.0015} & 0.6481 & 0.1494 & 0.0818 & 0.1065 & 0.9651 & 0.6021 & 0.4369 \\
 & $\bar{\rho} \uparrow$ & \textbf{1.0000} & 0.9598 & 0.9795 & 0.9765 & 0.9822 & 0.5576 & 0.6116 & 0.7161 \\
\midrule
\multirow{3}{*}{Support2} & Imp\% $\uparrow$ & 67.1\% & 78.0\% & 16.6\% & 66.4\% & 98.7\% & 35.8\% & \textbf{99.1\%} & 79.1\% \\
 & $\mathrm{SWD} \downarrow$ & \textbf{0.0134} & 0.1205 & 0.0175 & 0.0357 & 0.5662 & 0.2930 & 0.7766 & 0.5300 \\
 & $\bar{\rho} \uparrow$ & \textbf{0.9993} & 0.9558 & 0.9985 & 0.9803 & 0.6236 & 0.5453 & 0.2645 & 0.4713 \\
\midrule
\multicolumn{10}{c}{\textit{Time-Series Datasets}} \\
\midrule
\multirow{3}{*}{Daily Climate} & Imp\% $\uparrow$ & 87.2\% & 21.3\% & -15.3\% & 1.4\% & 91.4\% & 70.1\% & \textbf{95.9\%} & 81.0\% \\
 & $\mathrm{SWD} \downarrow$ & 0.0330 & 0.0704 & 0.3547 & $\boldsymbol{3.1\!\times\!10^{-16}}$ & 0.2970 & 0.4255 & 0.5477 & 0.4335 \\
 & $\bar{\rho} \uparrow$ & 0.9898 & 0.9958 & 0.9528 & \textbf{1.0000} & 0.8688 & 0.7464 & 0.6907 & 0.8020 \\
\midrule
\multirow{3}{*}{ECL} & Imp\% $\uparrow$ & \textbf{98.4\%} & 65.1\% & -5.1\% & 23.0\% & 23.5\% & 34.0\% & 43.9\% & 26.1\% \\
 & $\mathrm{SWD} \downarrow$ & \textbf{0.0120} & 0.1160 & 0.0437 & 0.0415 & 0.2910 & 0.2529 & 0.5820 & 0.1237 \\
 & $\bar{\rho} \uparrow$ & \textbf{0.9995} & 0.9506 & 0.9904 & 0.9746 & 0.9197 & 0.9761 & 0.4855 & 0.9619 \\
\midrule
\multirow{3}{*}{ETT} & Imp\% $\uparrow$ & 90.9\% & 15.2\% & 0.8\% & 8.2\% & 89.9\% & 76.3\% & \textbf{98.9\%} & 91.6\% \\
 & $\mathrm{SWD} \downarrow$ & \textbf{0.0073} & 0.0619 & 0.0325 & 0.0183 & 0.1634 & 0.1061 & 0.2947 & 0.0708 \\
 & $\bar{\rho} \uparrow$ & 0.9983 & 0.9946 & \textbf{0.9989} & 0.9970 & 0.9271 & 0.9562 & 0.7872 & 0.9689 \\
\midrule
\multirow{3}{*}{Microsoft Stock} & Imp\% $\uparrow$ & \textbf{97.6\%} & 47.7\% & 16.8\% & 21.4\% & 25.0\% & 28.4\% & 38.4\% & 30.4\% \\
 & $\mathrm{SWD} \downarrow$ & 0.1288 & 0.0949 & 0.0472 & \textbf{0.0169} & 0.1040 & 0.6485 & 0.1733 & 0.0741 \\
 & $\bar{\rho} \uparrow$ & 0.9583 & 0.9954 & \textbf{0.9980} & 0.9968 & 0.9739 & 0.7915 & 0.9398 & 0.9791 \\
\midrule
\multirow{3}{*}{WTH} & Imp\% $\uparrow$ & 86.9\% & 16.1\% & 3.7\% & 1.1\% & 49.1\% & 97.2\% & \textbf{100.0\%} & 94.7\% \\
 & $\mathrm{SWD} \downarrow$ & 0.0375 & 0.0633 & \textbf{0.0146} & 0.0550 & 0.1807 & 0.0646 & 0.2202 & 0.0941 \\
 & $\bar{\rho} \uparrow$ & 0.9820 & 0.9946 & \textbf{0.9989} & 0.9796 & 0.9142 & 0.9521 & 0.8574 & 0.9480 \\
\bottomrule
\end{tabular}
}
\end{table*}

DenoGrad is highly competitive in raw predictive gain, with notable improvements such as $98.4\%$ on ECL and $97.6\%$ on Microsoft Stock, but its main advantage appears in data preservation. To validate this rigorously across datasets and methods without normality assumptions, we applied the non-parametric Friedman test followed by the Nemenyi post-hoc test at $\alpha=0.05$. Following Demšar \cite{demsar2006statistical}, this protocol is the standard for multi-algorithm, multi-dataset benchmarking because it compares ranking consistency rather than outlier-sensitive means.

\begin{figure}
    \centering
    \includegraphics[width=0.85\textwidth]{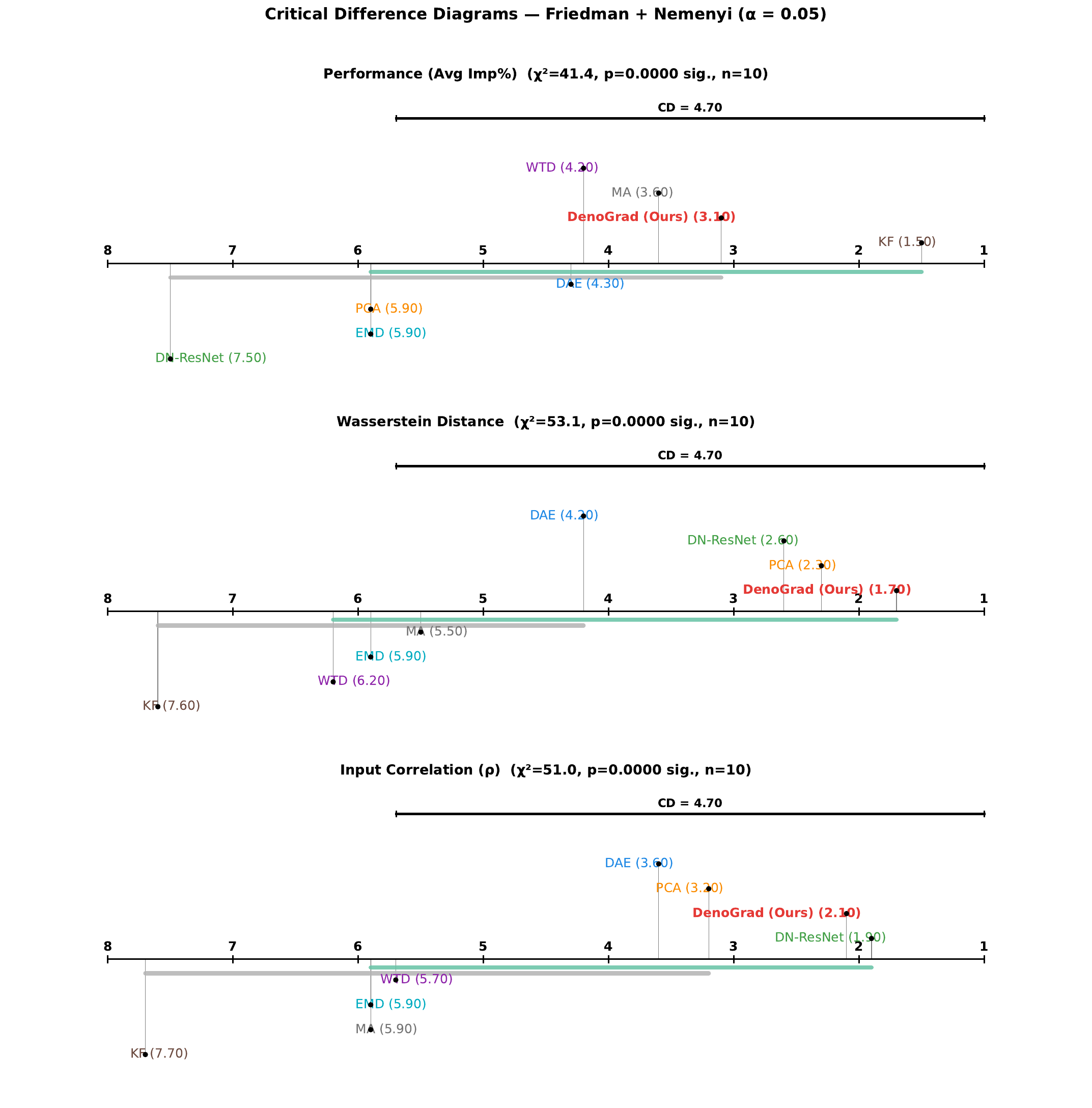}
    \caption{CD diagrams for the Nemenyi post-hoc test ($\alpha = 0.05$). Lower average rank (positioned further to the right) indicates better predictive performance or lower distribution distortion. Groups of regressors not significantly different are connected by a horizontal bar.}
    \label{fig:nemenyi_cd}
\end{figure}

The resulting Critical Difference diagrams in Figure \ref{fig:nemenyi_cd} visualize these rankings. DenoGrad achieves the best global average placement, leading in data integrity with mean ranks of $1.70$ (SWD) and $2.10$ ($\bar{\rho}$), while also reaching a top-tier rank of $3.10$ in predictive performance. Horizontal bars connect methods without statistically significant differences. DenoGrad consistently appears at the front of top-performing clusters, indicating stable performance across denoising scenarios and data regimes.

\begin{figure}
    \centering
    \begin{subfigure}{0.48\textwidth}
        \centering
        \includegraphics[width=\textwidth]{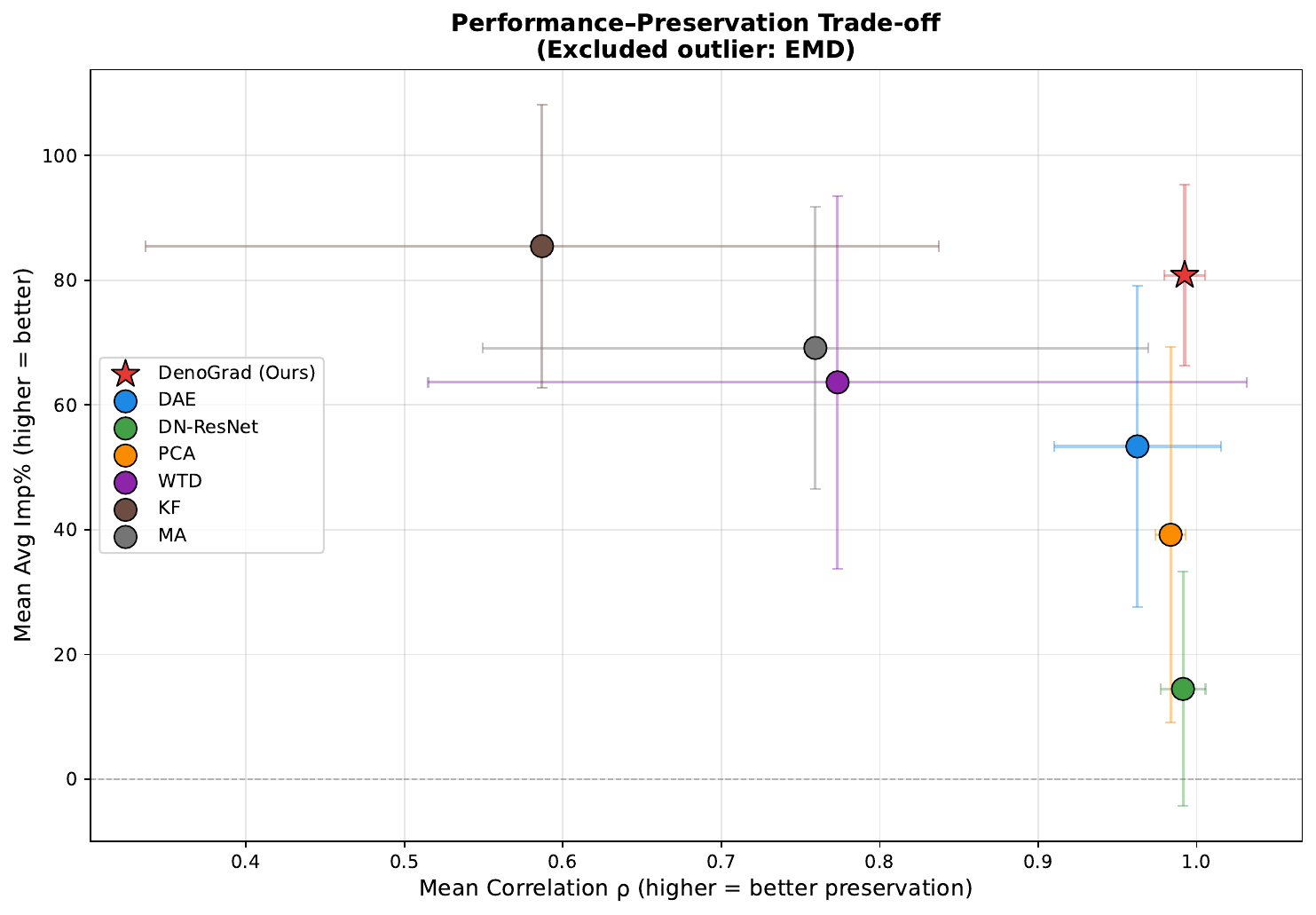}
        \caption{Pareto $\bar{\rho}$ x \%Improvement (top-right better)}
        \label{fig:pareto_corr}
    \end{subfigure}
    \hfill
    \begin{subfigure}{0.48\textwidth}
        \centering
        \includegraphics[width=\textwidth]{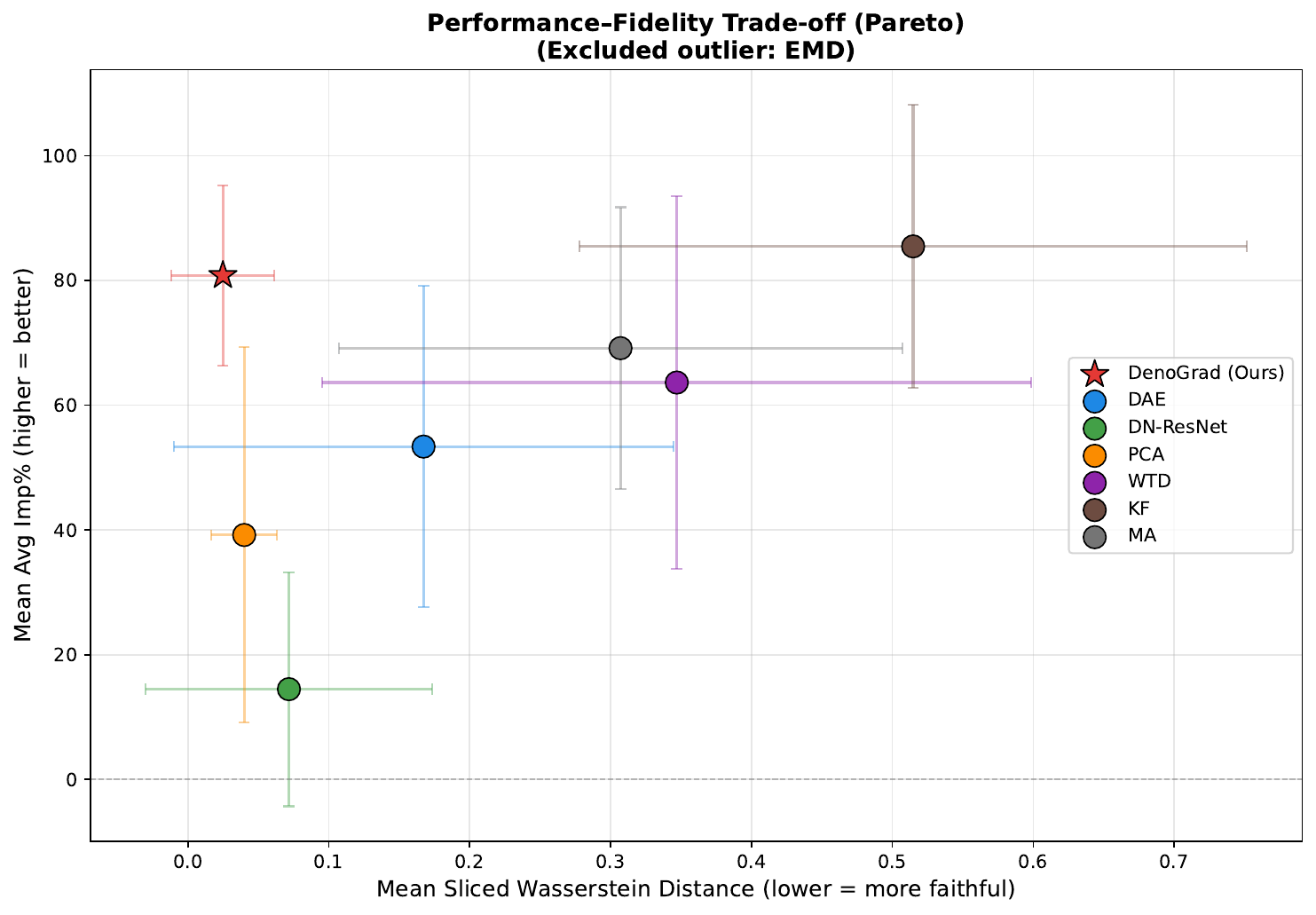}
        \caption{Pareto SWD x \%Improvement (top-left better)}
        \label{fig:pareto_swd}
    \end{subfigure}
    \caption{Pareto front plots illustrating the optimal trade-off between predictive improvement (y-axis) and correlation consistency (left) or data fidelity (right) achieved by DenoGrad.}
    \label{fig:pareto_tradeoff}
\end{figure}

Figure \ref{fig:pareto_tradeoff} further shows the performance-fidelity landscape. DenoGrad lies in the optimal region of both subplots: top-right in \textit{$\bar{\rho}$ vs.\ Imp\%} (high correlation preservation and high improvement) and top-left in \textit{SWD vs.\ Imp\%} (low SWD and high improvement). This indicates a favorable balance between aggressive noise reduction and structural preservation, consistent with Figures \ref{fig:pareto_swd} and \ref{fig:pareto_corr}, and in contrast to methods such as EMD or KF, which show stronger topological distortion.

\subsection{Ablation Study: Joint Feature-Target Optimization}
\label{sec:results;subsec:ablation_denoise_y}

Having established the global competitiveness of DenoGrad against external baselines, we now analyze one of its core design decisions: the joint optimization of both input features $X$ and the target variable $y$. In continuous regression and forecasting, $y$ is an intrinsic component of the data manifold rather than a discrete class label. Keeping $y$ frozen in a noisy state effectively forces the gradients $\nabla_X\mathcal{L}$ to align features with a corrupted anchor, systematically biasing the feature corrections. Jointly updating $X$ and $y$ resolves this by allowing each $(x,y)$ pair to move coherently toward a self-consistent equilibrium. Strictly speaking, this joint displacement does not guarantee the deterministic recovery of the exact latent ground truth. Instead, it functions as a highly effective, consistency-driven topological adjustment: both variables are iteratively shifted toward regions of lower model-induced discrepancy. Consequently, DenoGrad acts as a robust noise-reduction mechanism that faithfully reconstructs the structural geometry of the manifold, avoiding the gradient bias inherent to feature-only approaches.

This joint mechanism is especially critical in time-series settings with sliding windows, where targets at one horizon are coupled to feature channels in adjacent windows. If $y$ remains uncorrected, this mismatch propagates across temporal contexts and amplifies local optimization errors. To validate this empirically, we run a controlled ablation comparing joint optimization ($X{+}y$) against a feature-only variant ($X$-only) on all ten benchmark datasets.

In this setup, we inject Gaussian noise with standard deviation $\sigma = 0.1$ into both features and targets. We keep the same backbone configuration as in the main benchmark: DNN for tabular data and LSTM for time series. To avoid backbone-dependent reporting bias, performance is reported as mean downstream MSE across each domain's benchmark regressors (Table \ref{tab:downstream_params}).

Table~\ref{tab:ablation_denoise_y} reports the ablation results. The joint variant reaches a mean improvement (Imp\%) of $49.17\%$ versus $14.41\%$ for feature-only denoising, a $3.4{\times}$ gain in average predictive improvement. The effect is systematic, also reflected in the median ($45.08\%$ vs. $6.30\%$). The largest gaps appear in time series: on ECL, joint optimization achieves $95.5\%$ versus $4.4\%$ for $X$-only; on ETT, $X$-only degrades performance ($-3.9\%$) while $X{+}y$ recovers $30.9\%$. This shows how an uncorrected target can misguide gradient-based corrections.

\begin{table}[htbp]
\centering
\caption{Ablation study of target denoising ($y$) under Gaussian corruption with $\sigma=0.1$. Results are shown for the two denoising variants $X$-only and $X{+}y$. The clean-reference performance is reported in $\mathrm{MSE}_{\mathrm{clean}}$, computed on the original standardized dataset without injected Gaussian noise and without denoising. Tabular datasets use a DNN backbone and time-series datasets use an LSTM backbone. $\mathrm{MSE}_{\mathrm{deno}}$ is the mean MSE across downstream benchmark models after denoising for each variant and domain (see Table \ref{tab:downstream_params}). $\mathrm{Imp}\%$ is computed against the noisy baseline, i.e., the same dataset with injected $\sigma=0.1$ noise and no denoising. Lower is better for $\mathrm{SWD}$ and $\mathrm{MSE}$, higher is better for $\rho_X$, $\rho_y$, and $\mathrm{Imp}\%$.}
\label{tab:ablation_denoise_y}
\resizebox{0.8\textwidth}{!}{%
\begin{tabular}{ll r r r r r r r}
\toprule
Dataset & Variant & $\mathrm{SWD}_X$\,$\downarrow$ & $\mathrm{SWD}_y$\,$\downarrow$ & $\rho_X$\,$\uparrow$ & $\rho_y$\,$\uparrow$ & $\mathrm{MSE}_{\mathrm{deno}}$\,$\downarrow$ & $\mathrm{MSE}_{\mathrm{clean}}$\,$\downarrow$ & $\mathrm{Imp}\%$\,$\uparrow$ \\
\midrule
    \multirow{2}{*}{Daily Climate} & X-only & 0.0179 & 0.0375 & 0.9955 & 0.9844 & 0.105581 & \multirow{2}{*}{0.102276} & 5.1 \\
     & X+y (Ours) & \textbf{0.0048} & \textbf{0.0073} & \textbf{1.0000} & \textbf{1.0000} & \textbf{0.061586} &  & \textbf{44.6} \\
\midrule
    \multirow{2}{*}{House Prices} & X-only & 0.0203 & \textbf{0.0000} & 0.9948 & \textbf{1.0000} & 0.092581 & \multirow{2}{*}{0.151234} & 42.9 \\
     & X+y (Ours) & \textbf{0.0055} & 0.0432 & \textbf{0.9994} & 0.9867 & \textbf{0.088286} &  & \textbf{45.5} \\
\midrule
    \multirow{2}{*}{Lattice Physics} & X-only & 0.0033 & \textbf{0.0000} & 0.9998 & \textbf{1.0000} & \textbf{0.140003} & \multirow{2}{*}{0.140412} & \textbf{7.5} \\
     & X+y (Ours) & \textbf{0.0024} & 0.0090 & \textbf{0.9999} & 0.9978 & 0.144152 &  & 4.8 \\
\midrule
    \multirow{2}{*}{Parkinsons} & X-only & 0.0153 & \textbf{0.0000} & 0.9974 & \textbf{1.0000} & 0.320866 & \multirow{2}{*}{0.304663} & 24.8 \\
     & X+y (Ours) & \textbf{0.0066} & 0.0425 & \textbf{0.9992} & 0.9918 & \textbf{0.315472} &  & \textbf{26.1} \\
\midrule
    \multirow{2}{*}{ECL} & X-only & 0.0023 & 0.0109 & \textbf{1.0000} & 0.9998 & 0.950549 & \multirow{2}{*}{0.995392} & 4.4 \\
     & X+y (Ours) & \textbf{0.0003} & \textbf{0.0009} & \textbf{1.0000} & \textbf{1.0000} & \textbf{0.044935} &  & \textbf{95.5} \\
\midrule
    \multirow{2}{*}{ETT} & X-only & 0.0070 & 0.0242 & 0.9978 & 0.9851 & 0.045148 & \multirow{2}{*}{0.029602} & -3.9 \\
     & X+y (Ours) & \textbf{0.0017} & \textbf{0.0061} & \textbf{1.0000} & \textbf{1.0000} & \textbf{0.030045} &  & \textbf{30.9} \\
\midrule
    \multirow{2}{*}{Microsoft Stock} & X-only & 0.0119 & 0.0154 & 0.9996 & 0.9995 & 1.253971 & \multirow{2}{*}{0.965944} & 1.2 \\
     & X+y (Ours) & \textbf{0.0006} & \textbf{0.0008} & \textbf{1.0000} & \textbf{1.0000} & \textbf{0.119085} &  & \textbf{90.6} \\
\midrule
    \multirow{2}{*}{RT IOT 2022} & X-only & 0.0051 & \textbf{0.0000} & 0.9998 & \textbf{1.0000} & 1.139518 & \multirow{2}{*}{0.156265} & 1.1 \\
     & X+y (Ours) & \textbf{0.0012} & 0.0673 & \textbf{1.0000} & 0.9977 & \textbf{0.822494} &  & \textbf{28.6} \\
\midrule
    \multirow{2}{*}{Support2} & X-only & \textbf{0.0137} & \textbf{0.0000} & 0.9980 & \textbf{1.0000} & 0.251664 & \multirow{2}{*}{0.403129} & 47.1 \\
     & X+y (Ours) & 0.0139 & 0.0933 & \textbf{0.9989} & 0.9782 & \textbf{0.250757} &  & \textbf{47.3} \\
\midrule
    \multirow{2}{*}{WTH} & X-only & 0.0163 & 0.0713 & 0.9990 & 0.9930 & 0.355343 & \multirow{2}{*}{0.400932} & 14.0 \\
     & X+y (Ours) & \textbf{0.0014} & \textbf{0.0041} & \textbf{1.0000} & \textbf{0.9999} & \textbf{0.091325} &  & \textbf{77.9} \\
\bottomrule
\end{tabular}}
\end{table}

This gap is also visible in Appendix Figure~\ref{fig:ablation_mse}, which tracks downstream MSE across datasets. The $X{+}y$ variant consistently yields the lowest error, often approaching or surpassing the clean-data baseline. This is plausible because real datasets contain latent aleatory noise that the nominal clean baseline still retains; DenoGrad refinement can reduce both injected synthetic noise and part of this latent measurement noise.

Per-model heatmaps in Appendix Figure~\ref{fig:ablation_heatmap} show a clear domain asymmetry. In tabular datasets, $X$-only denoising already gives mostly positive gains, as $y$ is less structurally entangled with feature channels. In time series, however, $X$-only often degrades performance, for example LSTM ($-36.5\%$ in Daily Climate) and OhShuLih ($-29.8\%$ in ETT), supporting the sliding-window coupling hypothesis. With joint $X{+}y$ optimization, the time-series block becomes uniformly positive, with most cell-level gains above $60\%$, indicating that joint updates resolve this coupling issue across architectures and backbones.

Appendix Figure~\ref{fig:ablation_swd} further confirms distributional fidelity: the joint variant attains lower $\mathrm{SWD}_X$ in $9$ of $10$ datasets. This suggests a synergistic effect where correcting anchor $y$ reduces gradient bias and improves feature-space fidelity.

For $\mathrm{SWD}_y$, time-series datasets show non-zero values under $X$-only denoising (e.g., $0.0713$ in WTH), because sliding-window overlap makes feature updates indirectly affect adjacent targets. Joint optimization controls this effect, keeping $\mathrm{SWD}_y < 0.10$ while maintaining $\rho_X > 0.99$ across all datasets.

To formalize these observations, we apply a one-sided Wilcoxon signed-rank test ($X$-only $<$ $X{+}y$) over the ten paired Imp\% values. 
The test yields $W=4.0$ and $p=0.0068$, significant at both $\alpha=0.05$ and $\alpha=0.01$, therefore the improvement of performing X+y instead of X-only is significative and favorable for X+y case.


In summary, the ablation strongly supports joint optimization. Simultaneously refining features and targets removes a systematic gradient bias that strongly penalizes time-series tasks, yielding a statistically significant $3.4{\times}$ increase in predictive improvement. Importantly, this gain is achieved without sacrificing structure: the joint variant improves feature-distribution fidelity and preserves near-perfect correlations, indicating that target adjustment is a key step for reconstructing the underlying manifold.

\subsection{Generalization Across Downstream Models}
\label{sec:results;subsec:generalization}

A critical limitation of many conventional preprocessing techniques is their erratic behavior depending on the regressor used subsequently. To verify the universality of our approach, we analyzed the performance gains across five distinct downstream architectures. 

Imp\% values are inherently relative to the baseline regressor: a modest gain can indicate a model already near its ceiling, while a larger gain can reflect a weaker baseline. Because all denoisers are evaluated under the same pipeline, their Imp\% scores remain directly comparable. Our analysis therefore focuses on whether DenoGrad achieves systematically higher and more consistent improvements across downstream architectures.

\begin{figure}
    \centering
    \begin{subfigure}{0.48\textwidth}
        \centering
        \includegraphics[width=\textwidth]{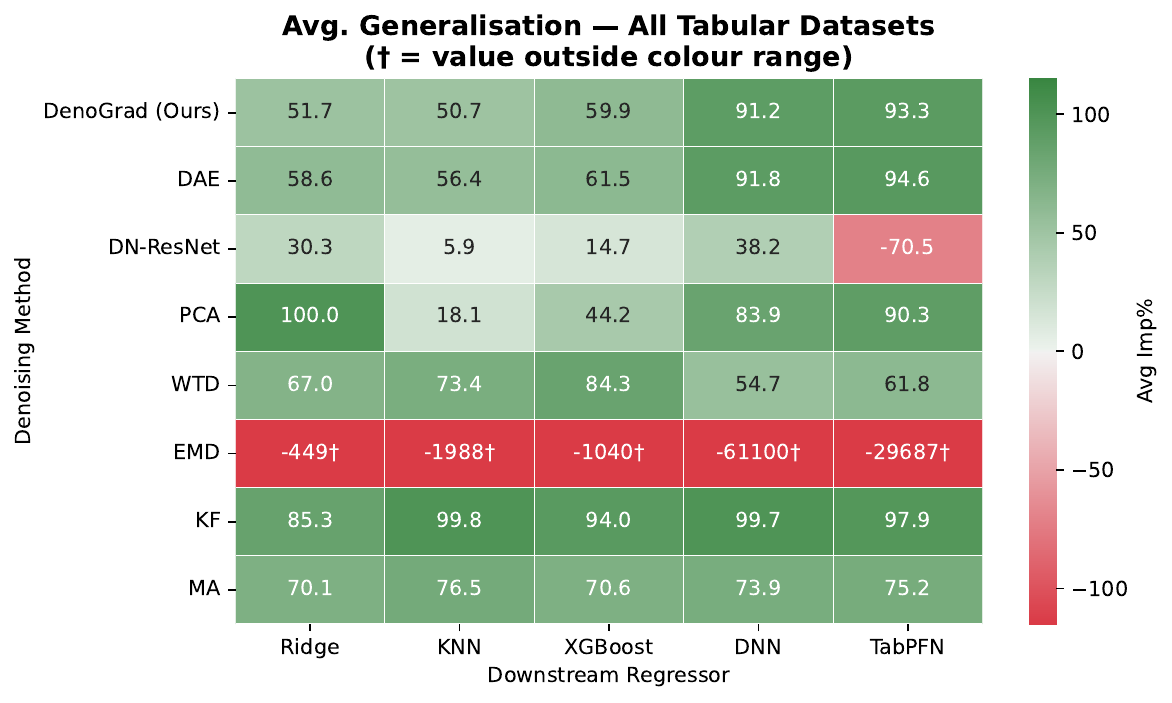}
        \caption{Tabular Domain (Averaged)}
        \label{fig:heatmap_tab}
    \end{subfigure}
    \hfill
    \begin{subfigure}{0.48\textwidth}
        \centering
        \includegraphics[width=\textwidth]{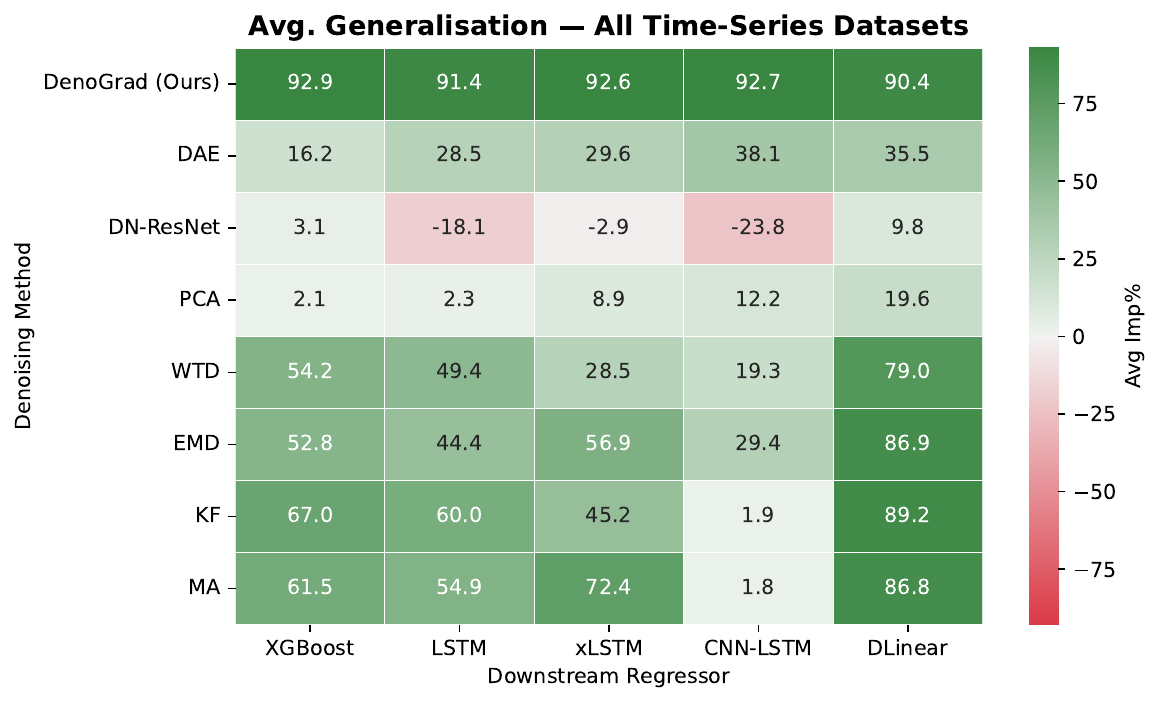}
        \caption{Time-Series Domain (Averaged)}
        \label{fig:heatmap_ts}
    \end{subfigure}
    \caption{Heatmaps illustrating the percentage improvement ($\mathrm{Imp}\%$) achieved by different denoisers across specific downstream regressors.}
    \label{fig:heatmap_downstream}
\end{figure}

Figure \ref{fig:heatmap_downstream} shows these interactions and a clear contrast in stability. Methods such as EMD or DN-ResNet fail on specific downstream models, including degradations with TabPFN or XGBoost, whereas DenoGrad remains uniformly positive and robust. In time series, Figure \ref{fig:heatmap_ts} identifies DenoGrad as the only method with consistent improvement above $90\%$ across LSTM, xLSTM, CNN-LSTM, DLinear, and XGBoost. This indicates that the manifold recovered by DenoGrad is broadly informative across inductive biases.

\subsection{Impact of the Backbone Architecture}
\label{sec:results;subsec:backbone}

As a model-independent framework, DenoGrad derives gradients from a DL backbone that need not be optimal but must capture the dominant data structure. We therefore performed an ablation to quantify how backbone choice affects denoising quality.

As shown in the radar charts (Appendix Figure \ref{fig:ablation_radar}), no backbone is universally optimal; performance follows architecture-dataset suitability. On tabular Parkinsons, a CNN backbone expands improvement across regressors and outperforms the baseline DenseNN. On time-series ECL, a standard LSTM provides weaker gradients, while replacing it with CNN-LSTM or xLSTM yields substantial gains (above $95\%$). These improvements are broadly distributed across downstream regressors rather than concentrated in isolated models, supporting DenoGrad's modular design and compatibility with future backbone advances.

\subsection{Hyperparameter Sensitivity and Convergence}
\label{sec:results;subsec:hyperparameters}

To characterize DenoGrad optimization dynamics, we performed a sensitivity analysis of NRR and gating threshold $\tau$ under different iteration limits. The goal is to clarify convergence-speed vs. stability trade-offs and guide hyperparameter choices that maximize gains at manageable cost.

Appendix Figure \ref{fig:hyper_sensitivity} aggregates results across datasets and shows that aggressive settings outperform conservative ones. In both tabular and time-series domains, higher reduction rates ($\eta \in \{0.01, 0.1\}$) move data quickly toward the target manifold, reaching peak performance around 200 iterations. Although these settings can later induce mild oscillations around gating boundaries, they escape deep corruption regions that lower rates cannot leave at same cost.

By contrast, the conservative setting $\eta=0.001$ is stable but inefficient. Even with a tenfold iteration budget increase, it does not match the asymptotic performance of more aggressive settings, suggesting insufficient momentum for effective landscape traversal.

Overall, we identify a practical configuration: moderate-to-high NRR ($\eta \approx 0.01$ to $0.1$) with early stopping around 200 iterations. For gating, the largest tested value ($\tau=0.1$) consistently performs best without clear saturation, making it a robust default. Users can increase $\tau$ further when stronger preservation of aleatory variability is desired.

\subsection{Robustness to Varying Noise Levels}
\label{sec:results;subsec:noise}

Real deployments often face variable and unpredictable corruption levels. To evaluate resilience, we varied synthetic Gaussian noise standard deviation on input features from $\sigma=0.0$ to $0.5$ ($0.05$, $0.1$, $0.2$, $0.5$) and tracked both performance and integrity metrics.

\begin{figure}
    \centering
    \includegraphics[width=0.8\textwidth]{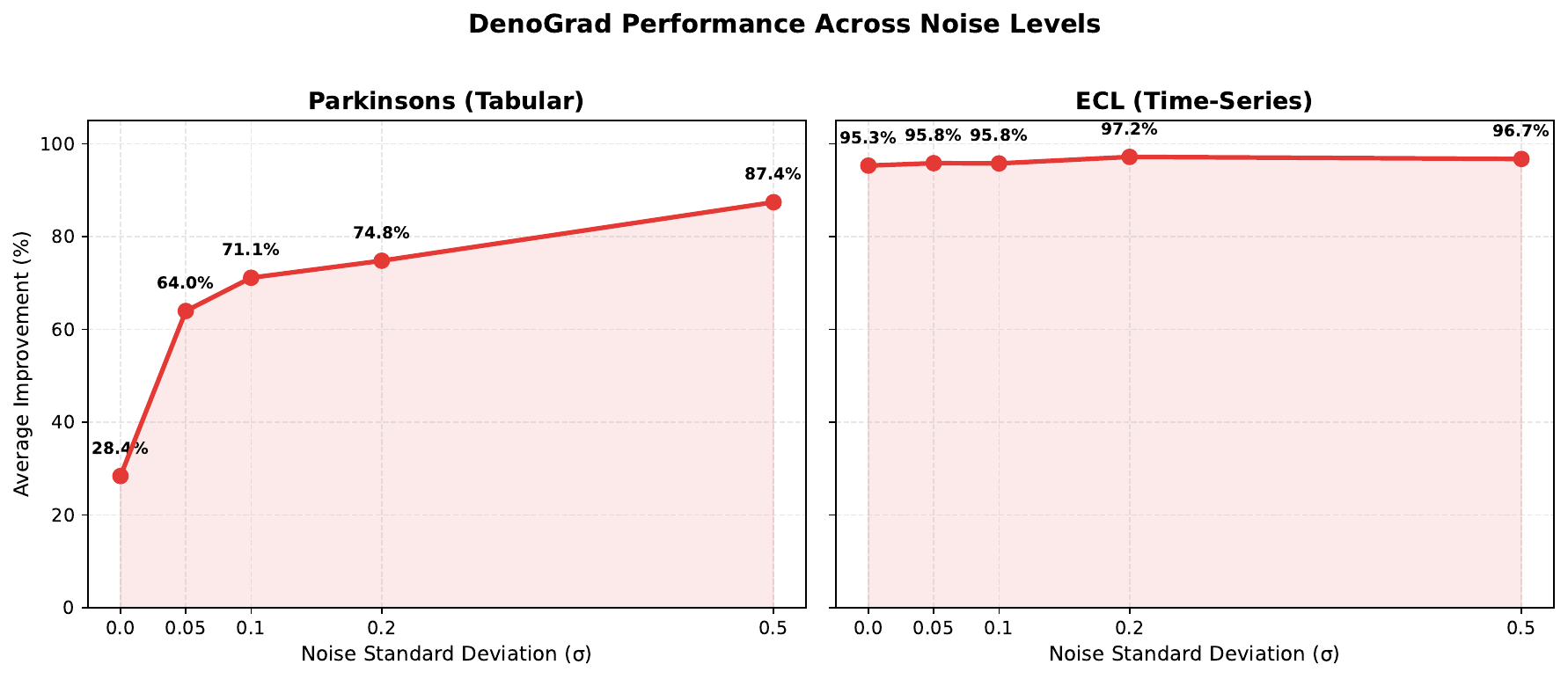}
    \caption{$Imp\%$ across varying standard deviations of injected noise ($\sigma$). DenoGrad maintains a near-constant high performance in Time-Series and scales proportionally with noise in Tabular data.}
    \label{fig:noise_robustness}
\end{figure}

As shown in Figure \ref{fig:noise_robustness}, DenoGrad exhibits modality-dependent but consistently effective behavior. For time series (ECL), performance is nearly invariant to noise intensity. Average predictive improvement remains between $95.3\%$ and $97.2\%$ across all tested levels. Appendix Figure \ref{fig:noise_heatmaps} confirms that this stability extends to all downstream regressors, which remain above $90\%$ improvement regardless of $\sigma$.

In the tabular domain (Parkinsons), relative improvement scales with noise intensity: average gain rises from $28.4\%$ to $87.4\%$ at $\sigma=0.5$. Per-model results show that robust models such as DenseNN benefit consistently, while sensitive architectures such as TabPFN dip at $\sigma=0.0$ but recover quickly, reaching $94.3\%$ as noise increases. This highlights DenoGrad's ability to recover heavily corrupted tabular samples.

Importantly, these gains do not sacrifice structural integrity. Appendix Figure \ref{fig:noise_quality} shows SWD remains tightly bounded across all $\sigma$, while $\bar{\rho}$ decreases only slightly to about $0.997$ in tabular data and stays near $0.9995$ in time series even at $\sigma=0.5$, avoiding the oversmoothing seen in competing methods.

A notable result is that DenoGrad improves performance even at $\sigma = 0.0$. This likely reflects suppression of latent aleatory measurement noise and dataset-level regularization that reduces inter-variable inconsistencies, improving generalization in nominally clean datasets.

\subsection{Computational Efficiency and Trade-offs}
\label{sec:results;subsec:efficiency}

DenoGrad cost is dominated by repeated forward/backward passes through the backbone. For dataset size $N$ and $E$ optimization iterations, complexity is $\mathcal{O}(E \cdot C_{model}(N))$, where $C_{model}$ is the cost of one forward-backward pass. In practice, $E$ is typically below 200, which keeps the method tractable for offline preprocessing.

Appendix Figure \ref{fig:scatter_efficiency} shows the runtime-quality trade-off using benchmark-averaged runtimes. DenoGrad is the most expensive method evaluated: about $77$s on tabular data and $543$s on time series. Statistical filters (MA, PCA, WTD) run in under $0.2$s, while learning-based baselines (DAE, DN-ResNet) range from $11$s to $41$s depending on modality.

These trade-offs must be interpreted jointly with integrity metrics. Fast statistical filters can match Imp\% in isolated cases, but, as shown in Section~\ref{sec:results;subsec:sota}, often with substantial topological distortion.

Although DenoGrad is the most computationally intensive method tested, the cost is often acceptable: denoising is typically a one-time offline preprocessing step. Online use may be possible depending on the backbone and the hyperparameters chosen.

\section{Discussion and Broader Implications}
\label{sec:discussion}

While DenoGrad shows strong empirical performance across datasets and noise regimes, deployment should be guided by a clear view of its mechanisms. This section discusses theoretical boundaries, practical configuration guidelines, and broader implications within machine learning.

\subsection{Assumptions and Limitations}
\label{sec:discussion;subsec:limitations}

DenoGrad effectiveness depends on several assumptions about both the predictive backbone and the noise profile.

First, denoising updates are derived entirely from the pretrained backbone loss landscape, so the method assumes that this backbone captures a sufficiently accurate approximation of the true signal (Section \ref{sec:gradient-based-denoiser}). If the backbone is strongly underfitted or biased, gradient directions may push samples toward a suboptimal manifold. This dependence is especially relevant under \textit{dataset shift}: if the target distribution differs substantially from the backbone training distribution, manifold misalignment can reduce structural fidelity. A natural mitigation path is to use stronger, more transferable foundational backbones.

Second, the method assumes corruption behaves approximately as zero-mean stochastic noise, consistent with the spectral-bias argument and with the Gaussian perturbations used in our controlled evaluation. Real data, however, may contain biased or structured components that violate this assumption. Although gains on nominally clean datasets suggest suppression of latent noise, this mechanism cannot yet be attributed conclusively.

More generally, DenoGrad has not been explicitly evaluated under strongly non-Gaussian, structured, or systematic noise. Under pronounced systematic bias, the backbone may encode that bias as signal, causing DenoGrad to preserve rather than correct it.

A final consideration is extreme outliers. Many gradient-based methods become unstable in these cases. DenoGrad mitigates this through batch-level gradient normalization: extreme outliers dominate magnitude, temporarily shrinking relative updates for in-distribution points and prioritizing correction of the outlier. As outlier error decreases, gradient balance is restored and convergence resumes. The threshold gating parameter $\tau$ stops updates for clean samples, improving stability overall. More iterations may be needed, especially with rare extreme anomalies. A lightweight pre-filtering or clipping step can significantly accelerate optimization.

\subsection{Practical Guidelines for Using DenoGrad}
\label{sec:discussion;subsec:guidelines}

The following recommendations are derived from our empirical analysis and are intended to support practical deployment of DenoGrad as an offline preprocessing step or an incremental refinement layer for streaming environments.

Regarding the backbone employed, it should demonstrate reasonable predictive performance on the task of interest. It need not be optimal, but must capture the dominant data structure. MLPs or CNNs are recommended for tabular data, and LSTM or CNN-LSTM for sequential data. Moderate-to-high values ($\eta \approx 0.01$ to $0.1$) achieve the best balance between escape from deep corruption and convergence stability. Use $\eta = 0.01$ as a robust default. Use smaller values only for light corrections in low-noise regimes. Values around $\tau = 0.1$ balance aggressive denoising with structural preservation. Increase $\tau$ to retain more aleatory variability. Cap iterations at 100 to 200 with early stopping when no sample exceeds $\tau$. Most gains are achieved within this budget.





\subsection{Implications for Data-Centric AI}
\label{sec:discussion;subsec:datacentric}

The results suggest implications beyond denoising alone. Data-Centric AI has highlighted that gains in data quality, labeling quality, and curation can exceed gains from architecture changes \cite{zha2025data}. In this context, DenoGrad shows how predictive models can be used not only to learn from data, but also to refine it.

Instead of treating denoising as purely statistical preprocessing, DenoGrad frames it as model-guided data refinement. By leveraging representations learned by a trained neural network, it iteratively moves samples toward regions more consistent with the learned manifold. As an inverse-adversarial mechanism, it uses the model itself as a semantic guide for improving dataset quality, connecting model-centric learning with data-centric optimization.

More broadly, DenoGrad emphasizes that model development and dataset engineering are complementary. Predictive models need not serve only as pipeline endpoints, but also be operational tools for improving data consistency and quality. Under this view, gradient-based dataset refinement becomes more than denoising. It becomes a concrete Data-Centric AI practice that reuses learned representations for curation.

\section{Conclusions and Future Work}
\label{sec:conclusions}

In this work, we introduced DenoGrad, a model-independent framework that reframes denoising as input optimization. By freezing a pretrained neural backbone and backpropagating corrections directly into feature and target space, DenoGrad corrects corrupted observations in both training and unseen data without requiring clean ground truth. This design reuses representational knowledge and spectral bias learned during training to guide noisy samples toward the learned manifold.

Our evaluation on ten real-world tabular and time-series datasets shows strong and consistent downstream gains. CD and Pareto analyses indicate a favorable balance between predictive improvement and topological preservation, with consistently low SWD and high $\bar{\rho}$ relative to competing methods. The refined manifold also transfers across diverse downstream regressors, from classical linear models to architectures such as TabPFN and xLSTM.

The joint-denoising ablation further shows that refining features and targets together outperforms feature-only refinement, supporting the core design choice. DenoGrad also behaves as an intrinsic regularizer, reducing latent aleatory noise and inter-variable inconsistencies, with gains even in nominally uncorrupted datasets.

These benefits are conditional. Performance depends on backbone quality and noise characteristics, and the method is mainly designed and validated for stochastic perturbations close to zero-mean Gaussian noise. Under strong systematic bias or substantial distribution shift, gradients may follow distortions encoded by the backbone rather than the true signal. These are key directions for future work.

More broadly, DenoGrad supports a data-centric view in which predictive models are not only learners but also tools for dataset refinement. Future extensions include adaptive NRR strategies for improved convergence stability, integration of large-scale foundational backbones to reduce task-specific dependence and improve robustness to dataset shift, and adaptation of gradient-based correction to classification, and large-scale curation pipelines.

\section*{Acknowledgments}
This publication is part of the Project “Ethical, Responsible and General Purpose Artificial Intelligence: Applications In Risk Scenarios” (IAFER) Exp.:TSI-100927-2023-1 funded through the Creation of university-industry research programs (Enia Programs), aimed at the research and development of artificial intelligence, for its dissemination and education within the framework of the Recovery, Transformation and Resilience Plan from the European Union Next Generation EU through the Ministry for Digital Transformation and the Civil Service. Ignacio Aguilera-Martos was supported by the Ministry of Science of Spain under the FPI programme PRE2021-100169.

\bibliographystyle{unsrt}  
\bibliography{references}  

\section*{Appendix}

\section{Additional Result Figures}

\begin{figure}
        \centering
        \includegraphics[width=0.8\textwidth]{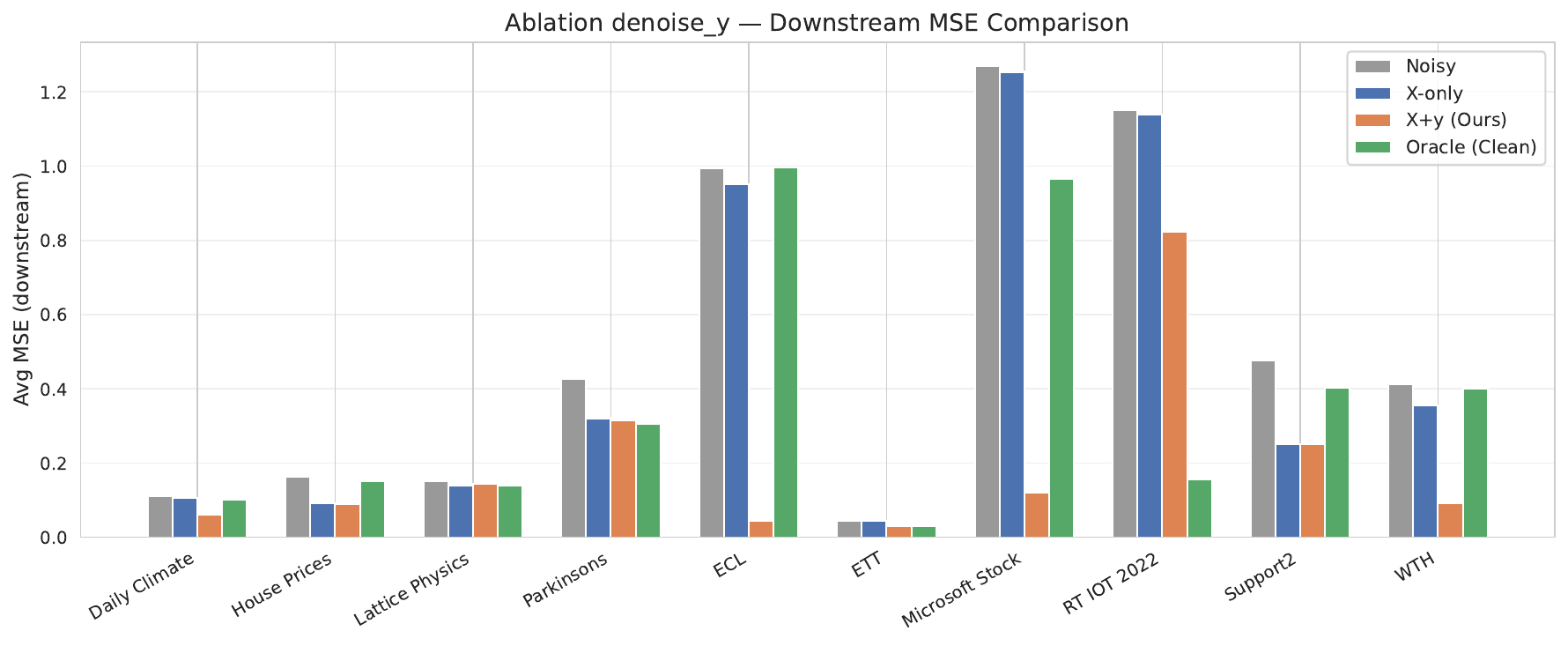}
        \caption{Downstream MSE comparison across all datasets. The $X{+}y$ joint optimization consistently achieves the lowest error, frequently matching or surpassing the clean-data baseline.}
        \label{fig:ablation_mse}
\end{figure}

\noindent\textbf{Figure~\ref{fig:ablation_mse}:} This complements Section~\ref{sec:results;subsec:ablation_denoise_y}. The conclusion is that joint optimization ($X{+}y$) reduces downstream error more consistently than $X$-only.

\begin{figure}
        \centering
        \includegraphics[width=0.9\textwidth]{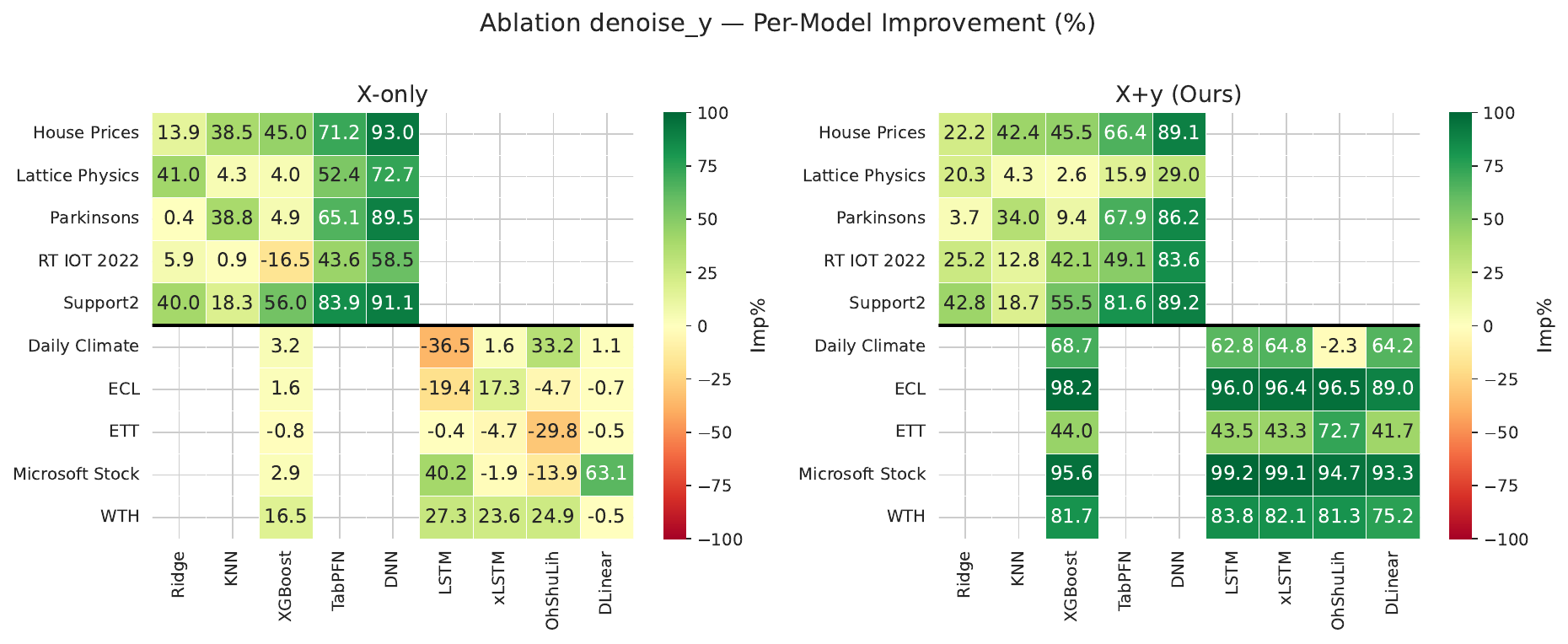}
        \caption{Per-model percentage improvement heatmaps for $X$-only (left) and $X{+}y$ (right). Static tabular datasets appear above the horizontal separator and time-series datasets below. Joint optimization turns the time-series block predominantly green, confirming that target denoising eliminates the gradient bias induced by the sliding-window coupling between features and targets.}
        \label{fig:ablation_heatmap}
\end{figure}

\noindent\textbf{Figure~\ref{fig:ablation_heatmap}:} This extends the ablation in Section~\ref{sec:results;subsec:ablation_denoise_y} to model-level detail. The key takeaway is that gains are broad across regressors, especially in time series when targets are denoised jointly.

\begin{figure}
        \centering
        \includegraphics[width=0.95\textwidth]{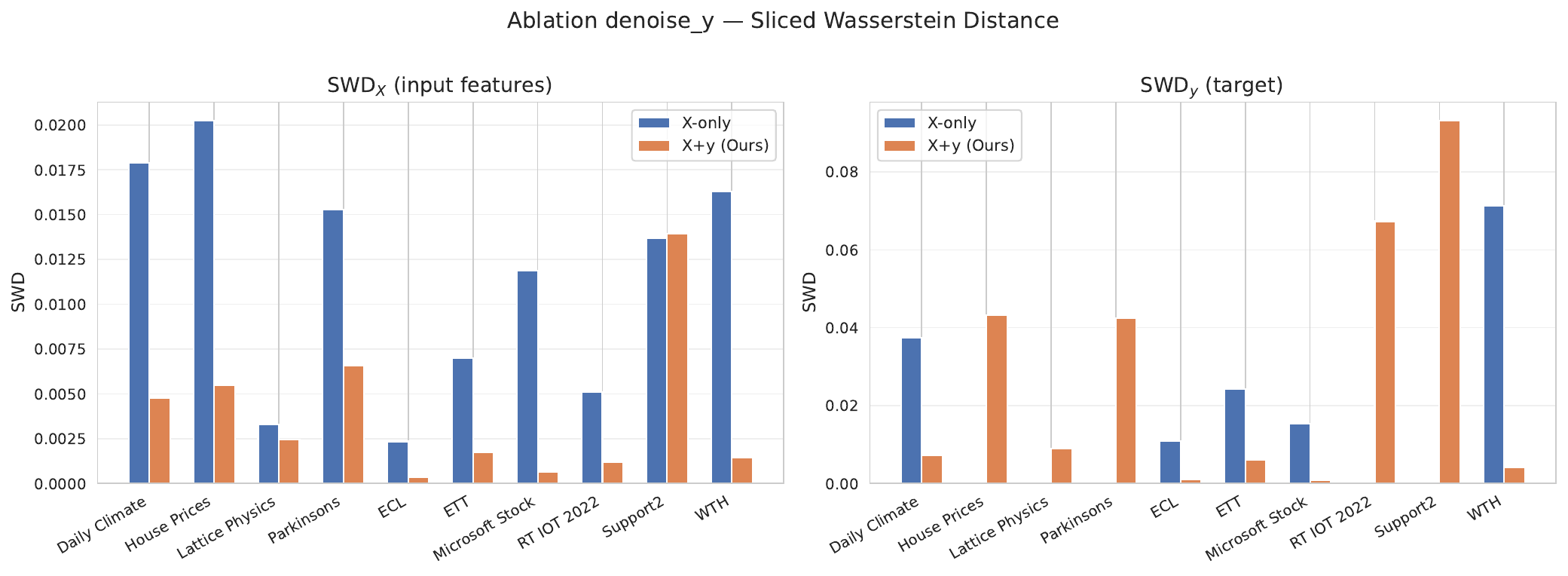}
        \caption{SWD for input features ($\mathrm{SWD}_X$, left) and target variable ($\mathrm{SWD}_y$, right). Joint optimization reduces $\mathrm{SWD}_X$ in the majority of datasets while maintaining controlled $\mathrm{SWD}_y$ values.}
        \label{fig:ablation_swd}
\end{figure}

\noindent\textbf{Figure~\ref{fig:ablation_swd}:} This supports the integrity analysis in Section~\ref{sec:results;subsec:ablation_denoise_y}. The conclusion is that joint optimization improves predictive performance while keeping distributional distortion controlled.

\begin{figure}
    \centering
    \includegraphics[width=0.75\textwidth]{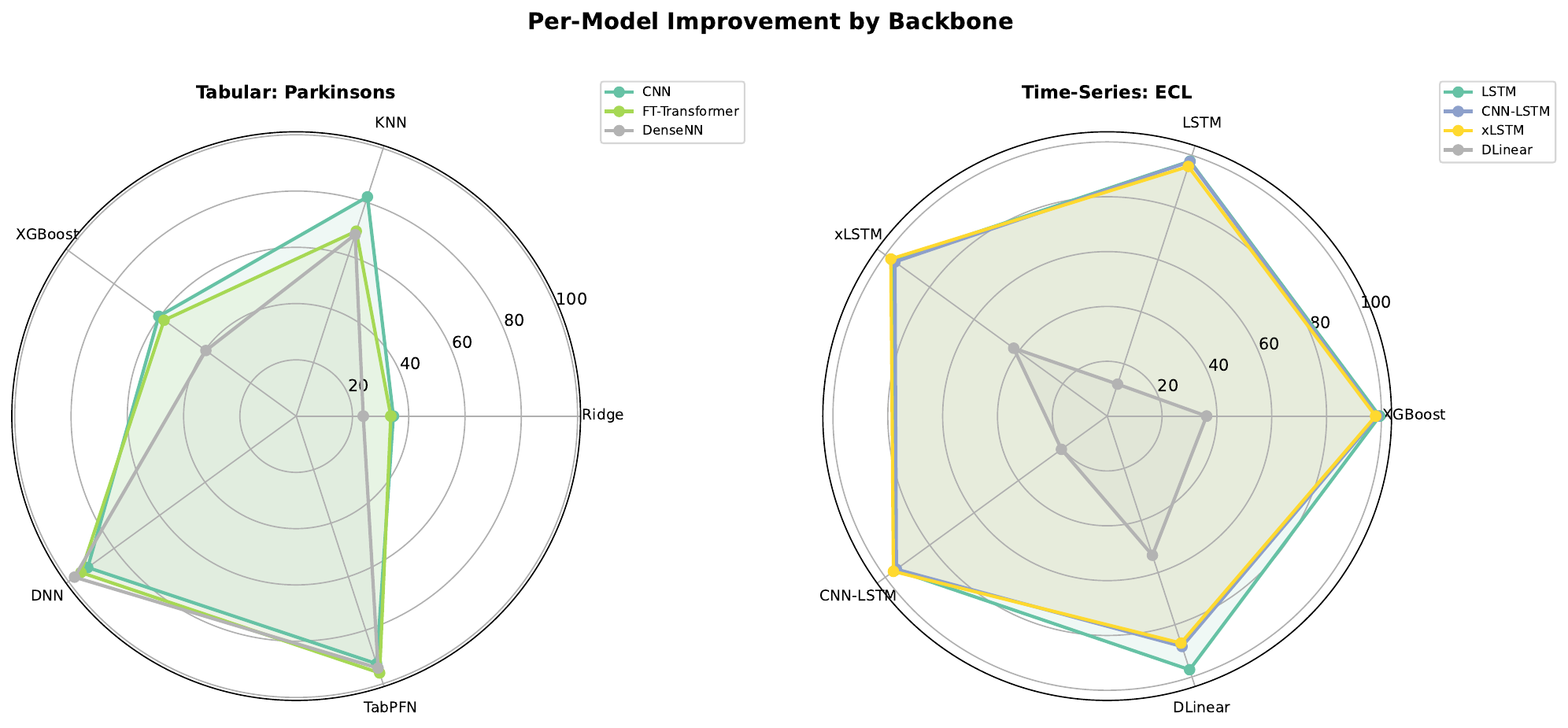}
    \caption{Model-specific improvement details, illustrating how the choice of backbone architecture affects individual downstream regressors uniformly.}
    \label{fig:ablation_radar}
\end{figure}

\noindent\textbf{Figure~\ref{fig:ablation_radar}:} This complements Section~\ref{sec:results;subsec:backbone}. The takeaway is that backbone choice affects gain magnitude, but improvements remain distributed across downstream models.

\begin{figure}
    \centering
    \begin{subfigure}{0.48\textwidth}
        \centering
        \includegraphics[width=\textwidth]{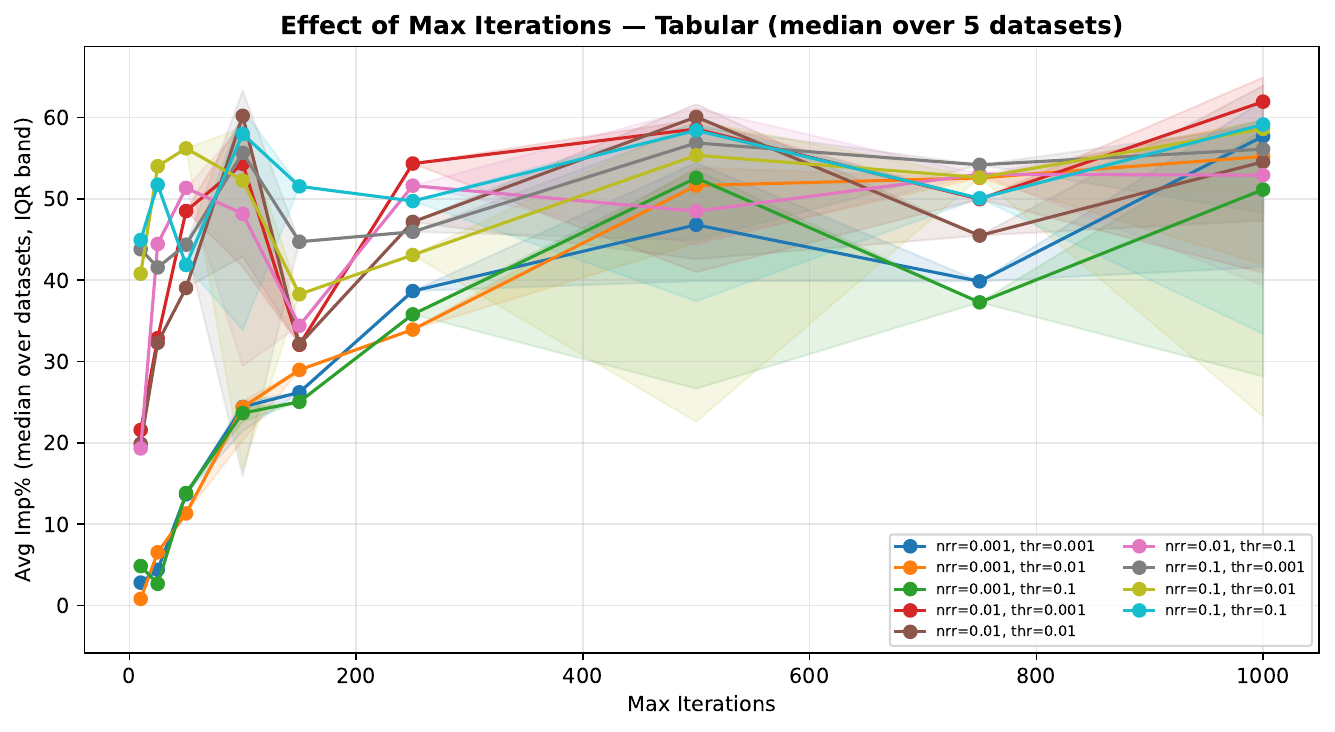}
        \caption{Tabular Domain (Median over 5 datasets)}
        \label{fig:hyper_tab}
    \end{subfigure}
    \hfill
    \begin{subfigure}{0.48\textwidth}
        \centering
        \includegraphics[width=\textwidth]{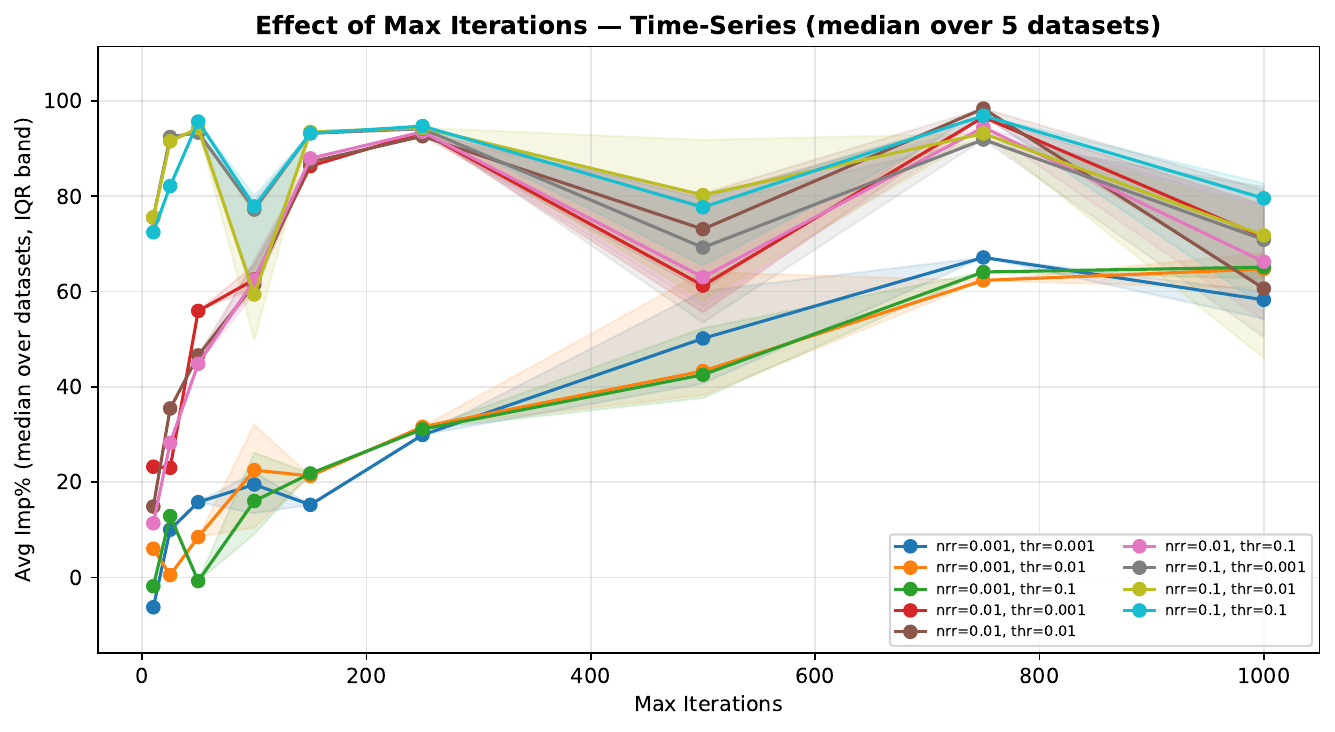}
        \caption{Time-Series Domain (Median over 5 datasets)}
        \label{fig:hyper_ts}
    \end{subfigure}
    \caption{Convergence analysis. The line plots illustrate the effect of maximum iterations on the average improvement (median $\pm$ IQR) across varying $\eta$ and threshold combinations.}
    \label{fig:hyper_sensitivity}
\end{figure}

\noindent\textbf{Figure~\ref{fig:hyper_sensitivity}:} This expands Section~\ref{sec:results;subsec:hyperparameters}. The main conclusion is that moderate-to-high $\eta$ values reach better performance faster, while very conservative settings converge slowly.

\begin{figure}
    \centering
    \includegraphics[width=0.8\textwidth]{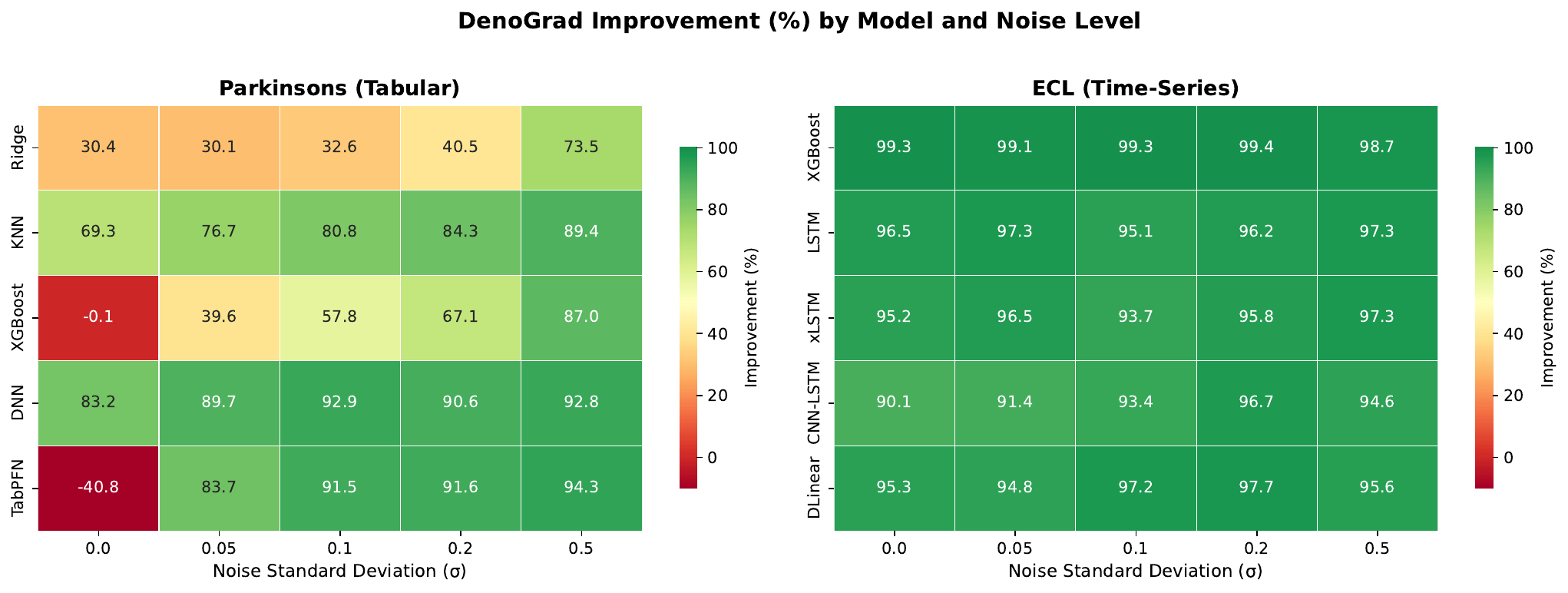}
    \caption{Heatmaps detailing the percentage improvement achieved by DenoGrad across individual downstream regressors at different noise levels.}
    \label{fig:noise_heatmaps}
\end{figure}

\noindent\textbf{Figure~\ref{fig:noise_heatmaps}:} This complements Section~\ref{sec:results;subsec:noise}. The takeaway is that DenoGrad remains effective across noise levels and across heterogeneous downstream regressors.

\begin{figure}
    \centering
    \includegraphics[width=0.8\textwidth]{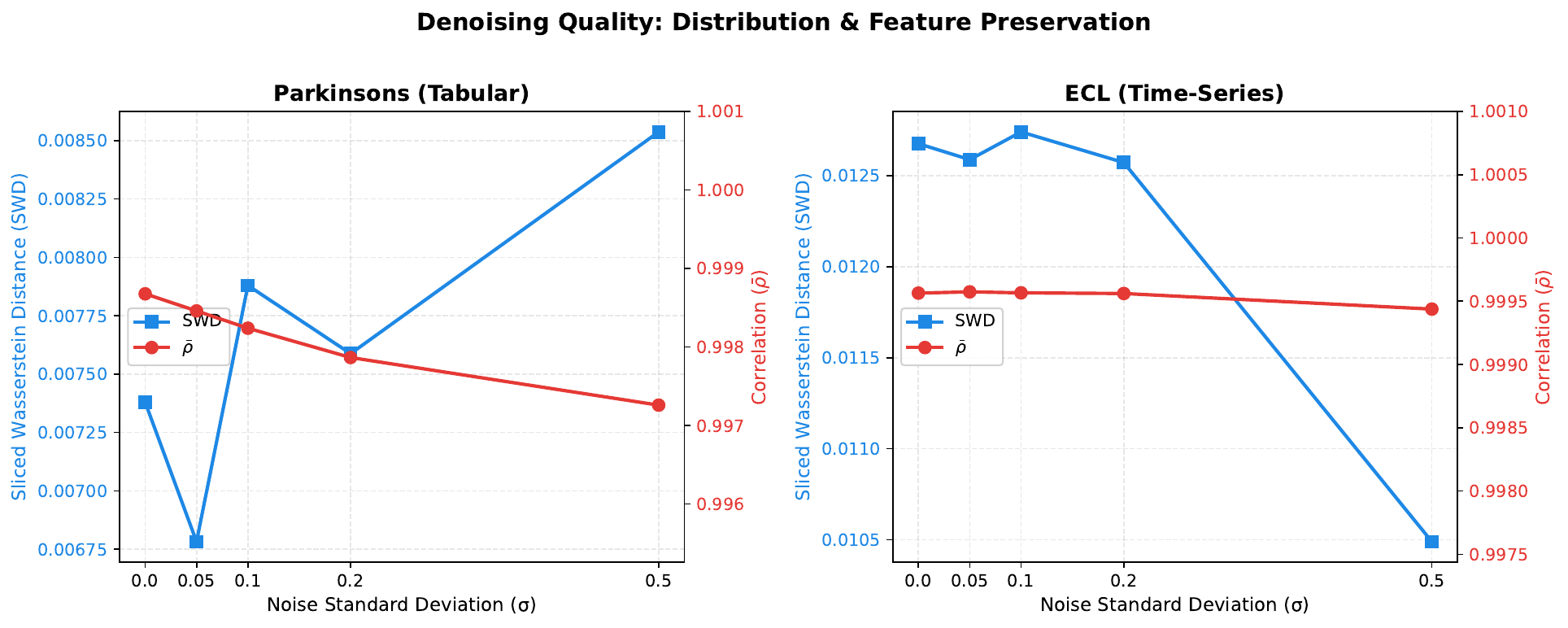}
    \caption{Data integrity preservation under varying noise conditions ($\sigma$). Feature correlation ($\bar{\rho}$, red line) remains highly stable near 1.0, while SWD (blue line) tracks the distributional shift.}
    \label{fig:noise_quality}
\end{figure}

\noindent\textbf{Figure~\ref{fig:noise_quality}:} This supports the integrity discussion in Section~\ref{sec:results;subsec:noise}. The conclusion is that quality gains do not come from aggressive oversmoothing, since correlation stays high while SWD stays bounded.

\begin{figure}
    \centering
    \includegraphics[width=0.85\textwidth]{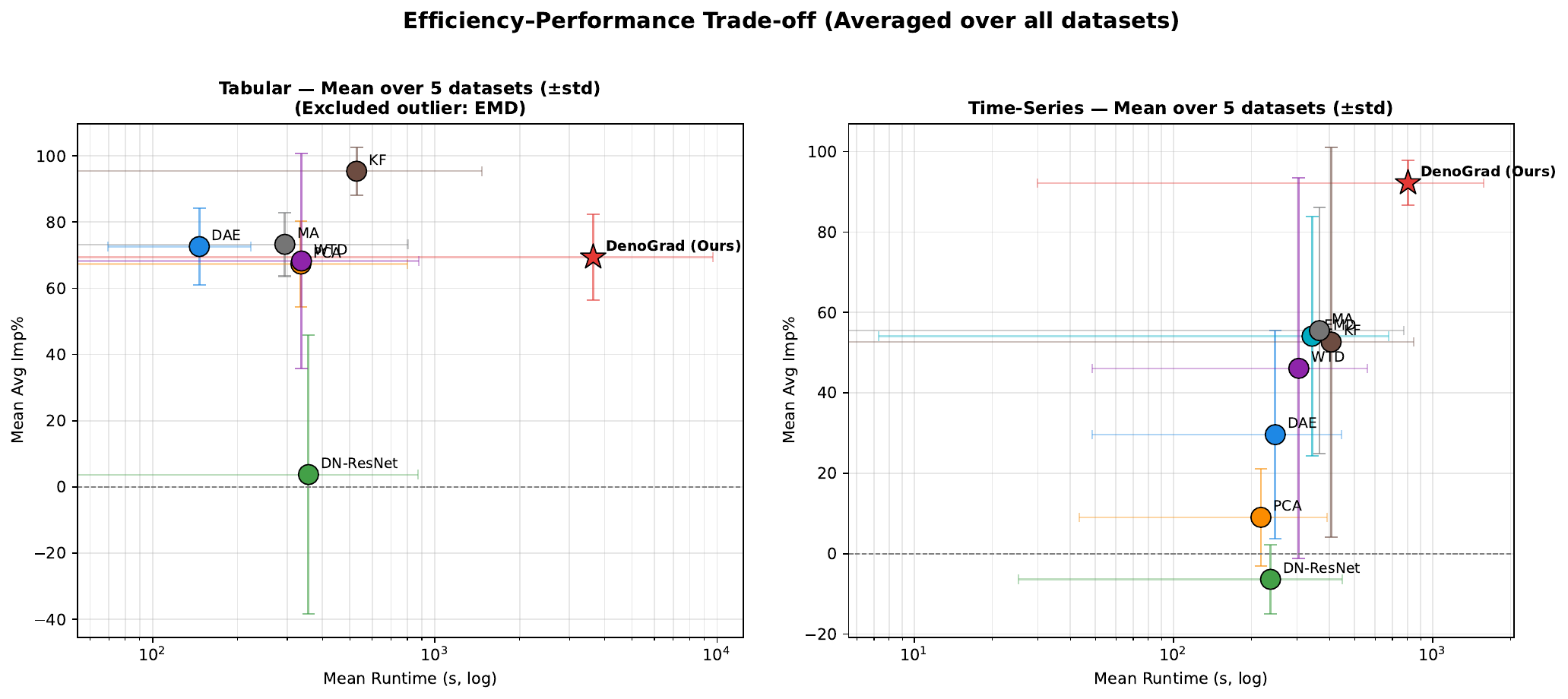}
    \caption{Efficiency-Performance landscape. Top-left better. The scatter plot illustrates the trade-off between computational cost (measured runtime in seconds on a logarithmic scale) and predictive gain (Average Improvement \%). Data points represent the aggregated mean performance across five tabular datasets (left) and five time-series datasets (right), with error bars denoting the standard deviation across independent runs.}
    \label{fig:scatter_efficiency}
\end{figure}

\noindent\textbf{Figure~\ref{fig:scatter_efficiency}:} This complements Section~\ref{sec:results;subsec:efficiency}. The key message is the expected trade-off: DenoGrad is slower, but offers a stronger quality-fidelity balance than faster baselines.

\end{document}